\documentclass[conference]{IEEEtran}

\usepackage{amsfonts, amsmath, amssymb}
\usepackage{booktabs}
\usepackage{float}
\usepackage[hidelinks]{hyperref}
\usepackage{multirow}
\usepackage{rotating}
\usepackage{scrextend}
\usepackage{tabularx}

\def\BibTeX{
    {\rm B\kern-0.05em{\sc i\kern-0.025em b}\kern-0.08em
    T\kern-0.1667em\lower0.7ex\hbox{E}\kern-0.125emX}
}

\begin{document}

\title{Benchmark of Data Preprocessing Methods for Imbalanced Classification}

\author{
    \IEEEauthorblockN{
        Radovan Halu\v{s}ka\IEEEauthorrefmark{1}\IEEEauthorrefmark{3},
        Jan Brabec\IEEEauthorrefmark{1}\IEEEauthorrefmark{2} and
        Tom\'{a}\v{s} Kom\'{a}rek\IEEEauthorrefmark{1}\IEEEauthorrefmark{2}
    }
    \IEEEauthorblockA{
        \IEEEauthorrefmark{1}Cisco Systems, Inc., Karlovo Namesti 10 Street, Prague, Czech Republic \\ \{rhaluska,janbrabe,tomkomar\}@cisco.com
    }
    \IEEEauthorblockA{
        \IEEEauthorrefmark{2}Czech Technical University in Prague, Faculty of Electrical Engineering, Czech Republic
    }
    \IEEEauthorblockA{
        \IEEEauthorrefmark{3}Faculty of Mathematics and Physics, Charles University, Prague, Czech Republic
    }
}

\IEEEoverridecommandlockouts
\IEEEpubid{
    \makebox[\columnwidth]{978-1-6654-8045-1/22/\$31.00~\copyright2022 IEEE \hfill}
    \hspace{\columnsep}
    \makebox[\columnwidth]{}
}

\maketitle
\IEEEpubidadjcol

\begin{abstract}
    Severe class imbalance is one of the main conditions that make machine learning in
    cybersecurity difficult. A variety of dataset preprocessing methods have been introduced over
    the years. These methods modify the training dataset by oversampling, undersampling or a
    combination of both to improve the predictive performance of classifiers trained on this
    dataset. Although these methods are used in cybersecurity occasionally, a comprehensive,
    unbiased benchmark comparing their performance over a variety of cybersecurity problems is
    missing. This paper presents a benchmark of 16 preprocessing methods on six cybersecurity
    datasets together with 17 public imbalanced datasets from other domains. We test the methods
    under multiple hyperparameter configurations and use an AutoML system to train classifiers on
    the preprocessed datasets, which reduces potential bias from specific hyperparameter or
    classifier choices. Special consideration is also given to evaluating the methods using
    appropriate performance measures that are good proxies for practical performance in real-world
    cybersecurity systems. The main findings of our study are: 1) Most of the time, a data
    preprocessing method that improves classification performance exists. 2) Baseline approach of
    doing nothing outperformed a large portion of methods in the benchmark. 3) Oversampling methods
    generally outperform undersampling methods. 4) The most significant performance gains are
    brought by the standard SMOTE algorithm and more complicated methods provide mainly incremental
    improvements at the cost of often worse computational performance.
\end{abstract}


\section{Introduction}

A classification problem is said to be class-imbalanced when the class prior probability of at
least one class, usually the class of interest, is significantly smaller than the prior probability
of some other class. Class-imbalanced problems occur across a variety of machine learning
application domains such as medicine~\cite{medical-imb-data},
finance~\cite{fraud-detection-imb-data, bank-fraud-imb-data}, astronomy~\cite{astronomy-imb-data}
and many others.

Specifically, in cybersecurity, virtually all of the frequently studied classification problems are
class-imbalanced (e.g. intrusion detection~\cite{bayesian-forests}, malware
detection~\cite{malware-detection}, phishing detection~\cite{phishing-detection}). Furthermore, the
class imbalance is frequently severe, with prior probabilities of the classes of interest being
$10^{-5}$ and lower~\cite{bayesian-forests} because severe malicious behaviour and attacks are
(thankfully) extremely rare. For example, in network telemetry, the majority of logs are related to
ordinary (benign) traffic, and only a tiny portion is related to malicious activities.
Interestingly, a class imbalance exists even in the already small portion of telemetry related to
malicious activities, as the prevalence of low-risk activities such as malicious advertising and
tracking is much greater than the prevalence of the most exciting threats with high severity (e.g.
remote access trojans, ransomware, APTs). The difficulties and the importance of the severe class
imbalance problem in cybersecurity were, to our knowledge, first mentioned by
Axelsson~\cite{axelsson} in 2000. Now, more than two decades later, a class imbalance is still
among the most critical factors that make machine learning in cybersecurity
difficult~\cite{dos-donts-cybersec, malware-detection-is-hard}.

While slight class imbalance does not usually pose a problem, once it reaches a certain degree,
machine learning classifiers without appropriate countermeasures cannot learn reliably from the
data~\cite{learning-from-imb-data}. In such cases, classifiers tend to become biased toward the
majority class and neglect the underrepresented one, resulting in a situation in which overall
accuracy is high due to the classifier predicting the majority class all of the time. However,
other, more relevant performance measures that reflect performance on all classes are poor.

Over the years, there has been a great deal of interest in the imbalanced classification problem.
Many different approaches were proposed spanning all the major stages of machine learning model
development. These stages are~\cite{ml-lifecycle}: 1) data management, 2) model learning and 3)
model verification. Approaches applied in the first stage are sometimes called \emph{data-level
methods}, while approaches applied in the second stage are called \emph{algorithm-level
methods}~\cite{johnson2019}. Multiple literature reviews~\cite{chawla2009, kotsiantis2006,
sokolova2009, learning-from-imb-data, johnson2019} summarising the concepts and popular approaches
in each stage have been published over time.

In this paper, we focus on data-level methods suitable for class-imbalanced learning. The idea
behind these methods is centred around modifying the distribution of the training dataset to make
it less imbalanced. This is, in principle, achieved via either oversampling the minority class or
undersampling the majority class. Many such methods have been published over the years, and
sometimes the rationale behind them is contradicting. The current situation concerning which
methods are worth using when and which are perhaps unnecessarily complex for little to no benefit
is unclear. In the worst case, this may lead to a promising, high-performing method being ignored
by the field in favour of a simpler or more traditional one. Our goal in this paper is to improve
understanding of strengths, weaknesses and various trade-offs (both predictive and computational)
between a range of the most well-known data-level methods.

To achieve this, we perform an extensive empirical benchmark of data-level methods on various
datasets spanning different application domains with special attention dedicated to the
cybersecurity domain. We aim to compare the methods objectively on as equal ground as possible,
which is helped by us not having any horses in the race. To the best of our knowledge, there does
not exist a more comprehensive benchmark of data oversampling and undersampling methods.  The
results help better navigate the problem landscape and select appropriate methods, hopefully
leading to improved predictive performance on various tasks in cybersecurity and other domains.

The rest of this paper is structured as follows. In the next section, we review the related work
and other benchmarks of data-level methods that were performed in the past.
Section~\ref{section:methods} introduces the methods that are included in the benchmark.
Section~\ref{section:setup} describes the setup of the experiment and its limitations.
Section~\ref{section:results} contains the experiment results, which are discussed in detail in
Section~\ref{section:discussion}. The main takeaways are summarised in
Section~\ref{section:conclusion}.

\section{Related Work}

Over the years, many data preprocessing methods suitable for class-imbalanced learning have been
published, but in comparison, only a relatively small number of benchmarks encompassing an
extensive range of both methods and datasets exist. Typically, every publication introducing a new
method includes experimental evaluation, but the scope of these experiments tends to be small. For
example, a paper introducing ADASYN~\cite{adasyn} contains experiments on five datasets and
compares the method only against SMOTE~\cite{smote} and plain decision tree baseline.

With that said, there already exist publications that focus mainly on comparing preprocessing
methods, but usually, they tend to focus only on oversampling methods. Most of these
studies~\cite{gosain2017, amin2016, barandela2004} are also performed on a relatively small number
of datasets. An exception is a study by Kovács~\cite{kovacs}, which is very extensive both in terms
of methods compared and datasets used. However, it focuses only on oversampling methods and also
does not contain experiments in the cybersecurity domain. Additionally, none of the studies above
performs as broad a search in hyperparameters and successive classifier models as we do.

In the cybersecurity domain, Wheelus et al.~\cite{wheelus2018} compared several preprocessing
methods on the UNSW-NB15~\cite{unsw} dataset. Bagui and Li~\cite{bagui2021} compared five
preprocessing methods on six network intrusion detection datasets and used a feed-forward neural
network with one hidden layer for classification. Furthermore, the most popular data preprocessing
methods are known and used in cybersecurity~\cite{ahsan2018, massaoudi2022, akash2022, soe2019,
azad2021}, but to our knowledge, a broader comparative study is missing.

Lastly, previous studies also summarise the results of individual methods into a single number.
Usually, this is the average rank or score the method achieved across all datasets. In this paper,
we provide rank distribution density plots instead of single numbers. These plots show a more
complete picture as the ranks tend to have a significant variance and overlap across the datasets.

\section{Benchmarked Methods}
\label{section:methods}

This section contains an overview of preprocessing methods used in the benchmark. For the sake of
space, we refrain from thorough explanations and refer to original publications.

\subsection{Oversampling Methods}

Oversampling methods constitute one possible approach to solving the imbalanced classification
problem. The main goal of oversampling methods is to modify the empirical distribution by
increasing the number of samples belonging to the minority class. The empirical distribution is
modified either by duplicating the existing samples or generating new artificial samples until the
desired imbalance ratio is reached.

The most straightforward approach is called \emph{Random Oversampling}, which, as its name
suggests, randomly duplicates already existing samples in the dataset.

One of the first and most widely used oversampling methods which produce synthetic data samples is
\emph{SMOTE}~\cite{smote}. It creates new synthetic examples on the line segments between existing
examples from the minority class. SMOTE, however, considers all of the minority samples to have the
same importance. It does not consider prior sample density and does not care about the
neighbourhood of a minority sample. Various enhancements have been proposed to aid these
shortcomings of the original SMOTE algorithm. We include four of those enhancements in our
benchmark, namely \emph{BorderlineSMOTE}~\cite{borderline-smote}, \emph{SVM
SMOTE}~\cite{svm-smote}, \emph{KMeansSMOTE}~\cite{kmeans-smote} and \emph{ADASYN}~\cite{adasyn}.

BorderlineSMOTE, as opposed to SMOTE, selects only minority samples with at least half of their
neighbours belonging to the majority class. The idea behind this approach is that minority samples
surrounded by more majority samples are close to the so-called decision boundary and are,
therefore, important in classification.

SVM SMOTE builds on the same idea but uses the SVM algorithm instead of the kNN algorithm to detect
minority samples near the decision boundary.

KMeansSMOTE tries to generate new artificial samples in regions where minority samples are sparse
and thus avoids further inflation of dense regions. It uses the KMeans clustering algorithm to
detect clusters containing more minority samples than majority samples. This avoids interpolation
between noisy minority samples. Subsequently, new samples are generated in each selected cluster
based on its density, i.e. more samples are generated in sparse clusters.

ADASYN differs from SMOTE by assigning weights to minority samples based on their difficulty in
learning. Difficulty in learning, in this case, means the portion of k-nearest neighbours that
belong to the opposite class. More synthetic data is generated in areas where it is hard to learn
minority samples, and less data is generated in other, easier-to-learn regions.

\subsection{Undersampling Methods}

Undersampling methods focus on the majority class, as opposed to the oversampling methods, to
address the issue of imbalanced classification. These methods reduce the number of samples in the
majority class to create a more balanced distribution of samples between classes. Most of the
undersampling methods discussed are so-called \emph{prototype selection} methods. Prototype
selection methods reduce the number of samples by removing unnecessary samples from the dataset and
using only a subset of the original data. The Cluster Centroids method is the only example of a
\emph{prototype generation} method used in the benchmark. Prototype generation methods reduce the
number of samples by generating new samples, e.g. centroids of clusters obtained by the KMeans
algorithm, instead of using a subset of the original ones.

Again, the simplest method based on random selection and removal of the majority samples is called
\emph{Random Undersampling}. The following several methods build on the kNN algorithm and modify it
to achieve slightly different results.

\emph{Condensed Nearest Neighbours - CNN}~\cite{cnn} reduces a potentially massive dataset into a
consistent dataset which, when used in the 1-NN rule, correctly classifies all of the examples from
the original dataset.

\emph{Edited Nearest Neighbours - ENN}~\cite{enn} classifies all samples in the class to
undersample by computing k-nearest neighbours for each on the whole original set. It then proceeds
to remove all such samples under consideration whose actual label does not match the label of most
of their neighbours.

\emph{Repeated Edited Nearest Neighbours}~\cite{repeated-enn} consists of repeating the previous
algorithm multiple times to reduce the number of majority samples even further.

\emph{All KNN}~\cite{repeated-enn} uses the same idea as the two previous preprocessing methods to
eliminate samples from the majority class when there is a label disagreement between a sample under
consideration and its k-nearest neighbours. However, instead of using a fixed number of neighbours
to check an agreement, it starts by looking at the single nearest neighbour, then two nearest
neighbours and so on, until it reaches k-nearest neighbours. A sample is kept in the majority class
only if its label agrees in all cases.

\emph{Near Miss}~\cite{near-miss} is a collection of three algorithms that use kNN to select
majority class samples to retain. Near Miss 1 selects majority samples that exhibit the smallest
average distance to $\mathrm{N}$ closest minority samples. In contrast, Near Miss 2 selects those
samples that exhibit the smallest average distance to $\mathrm{N}$ furthest minority samples. Near
Miss 3 selects a given number of closest majority samples for each minority sample.

\emph{Tomek Links}~\cite{tomek-links} is a data-cleaning technique used to clean the area near the
decision boundary by removing noisy samples called Tomek Links. A pair of points
$\mathbf{x}_i~\in~S_{min}$, $\mathbf{x}_j~\in~S_{maj}$ belonging to the opposite classes form a
Tomek Link if there does not exist a sample $\mathbf{x}_k$ such that $\mathrm{d}(\mathbf{x}_i,
\mathbf{x}_k) \leq \mathrm{d}(\mathbf{x}_i, \mathbf{x}_j)$ or $\mathrm{d}(\mathbf{x}_j,
\mathbf{x}_k) \leq \mathrm{d}(\mathbf{x}_i, \mathbf{x}_j)$.

\emph{One Sided Selection - OSS}~\cite{one-sided-selection} divides majority samples into noise
samples, borderline samples, redundant samples and safe samples. OSS detects and removes redundant
samples using CNN. Afterwards, Tomek Links removes noise and borderline samples, leaving us with
only safe samples which are kept.

\emph{Neighbourhood Cleaning Rule - NCL}~\cite{ncl} tries to mitigate one major drawback of the OSS
method by replacing CNN with ENN, as CNN tends to retain noisy samples in the training
set~\cite{ncl}. Additionally, NCL cleans the neighbourhoods of minority samples. For each minority
sample, it computes its three-nearest neighbours. If those neighbours misclassify the minority
sample under consideration, it removes the neighbours of that sample that belong to the majority
class.

\emph{Cluster Centroids}~\cite{cluster-centroids} is our last method in the benchmark and is the
only method representing prototype generation undersampling methods. It uses the KMeans algorithm
to cluster majority class samples into clusters and outputs clusters' centroids as new majority
samples.

\section{Experiment Setup}
\label{section:setup}

\begin{figure}
    \centering
    \includegraphics[width=\linewidth]{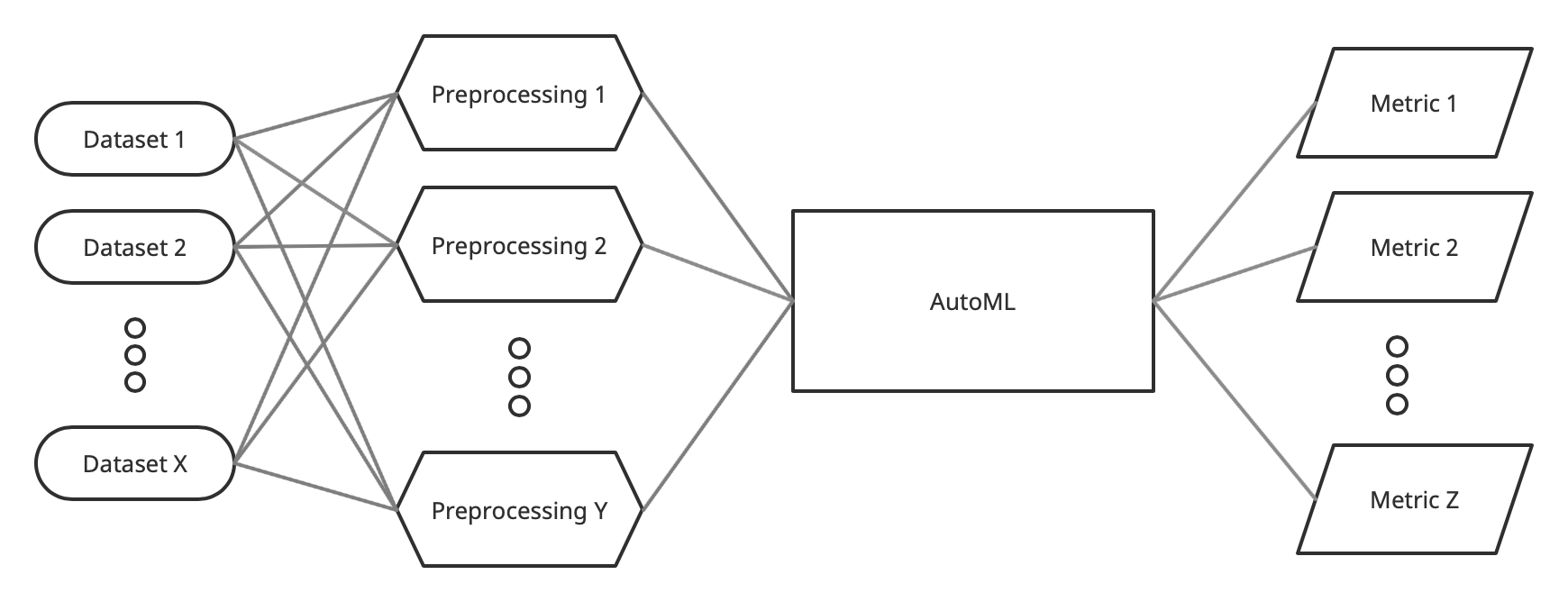}
    \caption{\textbf{High-level architecture of the benchmarking framework.}}
    \label{figure:framework}
\end{figure}

We built a benchmark framework to efficiently and robustly conduct experiments with many different
preprocessing methods over many datasets reporting as many evaluation metrics as supplied. The core
idea of the framework is depicted in Figure~\ref{figure:framework}. Each run combines a dataset, a
preprocessing method, and an instantiation of its hyperparameters found using a grid search. In
each run, a preprocessing method is applied to the training part of a dataset, yielding a new
resampled training set, which is then passed to the AutoML component of the framework. We use a
state-of-the-art AutoML framework Auto-Sklearn~\cite{auto-sklearn-1.0} for selecting, training and
tuning a classifier suitable for a given dataset. We provide more details about Auto-Sklearn in
Section~\ref{subsubsection:auto-sklearn}. Once a classifier has been trained, we perform
predictions using unseen examples from the test set and report evaluation scores achieved.

\subsection{Benchmark Setup}

We ran a benchmark covering 16 preprocessing methods discussed in Section~\ref{section:methods} and
one no-op baseline method. We covered several possible hyperparameter configurations for each
method shown in Table~\ref{table:configs}. All implementations of preprocessing methods used in the
benchmark originate from the \emph{Imbalanced Learn} library~\cite{imblearn}.

Every preprocessing method was run on 23 public and proprietary datasets shown in
Table~\ref{table:datasets}. Non-cybersecurity public datasets were downloaded from
OpenML~\cite{openml}. We chose datasets carefully based on multiple criteria such as dataset size,
amount of missing values and imbalance ratio. We required every dataset from OpenML to be binary
and to have at least 5000 samples; at most 20\% of samples could have missing values, and the
minimal imbalance ratio had to be 1:10. Although we focus only on binary classification, imbalanced
datasets occur in the multi-class environment as well. However, for the sake of simplicity and
consistency with other authors and publications, we focus only on the binary case. The
generalisation to the multi-class setting can be easily achieved by employing one-vs-one or
one-vs-rest schemes to preprocessing methods and micro and macro averaging to evaluation metrics.
We used 75\% of data samples from each dataset as a training set and the remaining 25\% as a
testing set. The split was done to retain the original imbalance in both sets.

We utilised Auto-Sklearn~\ref{subsubsection:auto-sklearn} to find, train and tune the
best-performing classifier on the training set using five-fold cross-validation as the validation
technique. Auto-Sklearn was set to optimise the ROC AUC score~\ref{subsubsection:roc-curve}. Each
run was given a total of 30 minutes for training on public datasets; a single machine learning
model had 10 minutes to finish training. Unsuccessful runs were not repeated. Due to their sizes,
it was sufficient to dedicate only five minutes to Auto-Sklearn on proprietary datasets, and no
repetitions were needed. We did not limit the time for preprocessing step in any way to obtain data
about the performance of preprocessing methods on datasets of different sizes.

\begin{table}[ht]
    \centering
    \begin{tabular}{lr}
        Method & Hyperparameter Configurations \\
        \midrule
        Baseline & 1 \\
        Random Oversampling & 2 \\
        SMOTE & 4 \\
        Borderline SMOTE & 16 \\
        SVM SMOTE & 8 \\
        KMeans SMOTE & 4 \\
        ADASYN & 4 \\
        Random Undersampling & 2 \\
        CNN & 2 \\
        ENN & 4 \\
        Repeated ENN & 4 \\
        All KNN & 4 \\
        Near Miss & 12 \\
        Tomek Links & 1 \\
        One-Sided Selection & 2 \\
        NCL & 8 \\
        Cluster Centroids & 4 \\
        \midrule
        $\Sigma$ & 82 \\
    \end{tabular}

    \vspace{4mm}

    \caption{
        \textbf{Hyperparameter Configurations for Preprocessing Methods.} The table shows the
        number of available hyperparameter configurations in the benchmark.
    }
    \label{table:configs}
\end{table}

\begin{table}
    \centering
    \begin{tabular}{lrrr}
        Name & Maj. Size & Min. Size & Imbalance \\
        \midrule
        Asteroid & 125,975 & 156 & 807.532 \\
        Credit Card Subset~\cite{credit-card} & 14,217 & 23 & 618.130 \\
        Credit Card~\cite{credit-card} & 284,315 & 492 & 577.876 \\
        PC2~\cite{pc2+mc1} & 5,566 & 23 & 242.000 \\
        MC1~\cite{pc2+mc1} & 9,398 & 68 & 138.206 \\
        Employee Turnover & 33,958 & 494 & 68.741 \\
        Satellite~\cite{satellite} & 5,025 & 75 & 67.000 \\
        BNG - Solar Flare & 648,320 & 15,232 & 42.563 \\
        Mammography & 10,923 & 260 & 42.012 \\
        Letter~\cite{lttr-dataset} & 19,187 & 813 & 23.600 \\
        Relevant Images & 129,149 & 5,582 & 23.137 \\
        Click Prediction V1 & 1,429,610 & 66,781 & 21.407 \\
        Click Prediction V2 & 142,949 & 6,690 & 21.368 \\
        Amazon Employee & 30,872 & 1,897 & 16.274 \\
        BNG - Sick & 938,761 & 61,239 & 15.329 \\
        Sylva Prior & 13,509 & 886 & 15.247 \\
        BNG - Spect & 915,437 & 84,563 & 10.826 \\
        \midrule
        CIC-IDS-2017~\cite{ids} & 227,132 & 5,565 & 40.814 \\
        UNSW-NB15~\cite{unsw} & 164,673 & 9,300 & 17.707 \\
        CIC-Evasive-PDF~\cite{pdf} & 4,468 & 555 & 8.050 \\
        Ember~\cite{ember} & 200,000 & 26,666 & 7.500 \\
        Graph - Embedding~\cite{dvorak2022} & 394 & 154 & 2.558 \\
        Graph - Raw~\cite{dvorak2022} & 394 & 154 & 2.558 \\
    \end{tabular}

    \vspace{4mm}

    \caption{
        \textbf{Datasets.} The table shows basic information about the datasets used in the
        benchmark. The upper part of the table shows publicly available non-cybersecurity datasets;
        the lower part shows cybersecurity datasets and two proprietary datasets concerning the
        classification of nodes in network graphs.
    }
    \label{table:datasets}
\end{table}

\subsubsection{AutoSklearn}
\label{subsubsection:auto-sklearn}

Auto-Sklearn~\cite{auto-sklearn-1.0} is a library for automated model selection and hyperparameter
tuning. Auto-Sklearn allows us to explore many models without introducing our own bias into the
process. We chose Auto-Sklearn for its significantly better performance than other competing AutoML
systems~\cite{auto-sklearn-1.0}. Although the second version of Auto-Sklearn, bringing substantial
advances~\cite{auto-sklearn-2.0}, has been available since 2020, we chose not to use it as it was
still in an experimental phase at the time of the experiments.

Auto-Sklearn extends existing AutoML architectures utilising the Bayesian optimiser by using
meta-learning and ensemble building to further boost the system's performance. We briefly explain
how each of the components works and provide comments in cases where we have needed to modify the
behaviour of Auto-Sklearn to allow complete control over the experiment.

Bayesian optimisation serves as an intelligent random search for hyperparameter tuning. It is a
powerful technique suitable for finding the extrema of objective functions expensive to evaluate,
such as tuning hyperparameters in a machine learning model, in as few sampling steps as
possible~\cite{bayesian-opt}. Bayesian optimisation fits a probabilistic model capturing a
relationship between hyperparameters and model performance. The probabilistic model suggests a
promising configuration of hyperparameters based on its current beliefs. It evaluates the model
using these hyperparameters and uses the results to update its beliefs in a loop. It can explore
new regions and exploit known regions to boost performance further~\cite{auto-sklearn-1.0}.

Meta-learning uses previous knowledge stored in a knowledge base consisting of pairs of dataset
characteristics and a machine learning model + hyperparameters exhibiting the best performance on
that dataset to suggest models that are likely to perform well on a new dataset. Characteristics
are viewed as vectors living in a normed meta-features vector space allowing us to use a notion of
distance between datasets to find similar ones and use models that performed well on those datasets
as a starting point for further tuning. Auto-Sklearn can quickly suggest configurations that are
likely to perform well on a new dataset to the Bayesian optimiser. Unfortunately, we encountered
various errors with meta-learning and could not get it to work correctly, so we disabled it
entirely for our experiment.

The data preprocessing step in Auto-Sklearn consists of missing values imputation, one-hot
encoding, data normalisation, scaling and centring. Feature preprocessing tries to create new
features using polynomial features or select a subset of features using PCA or ICA. Auto-Sklearn
also sometimes chooses to do the balancing. Balancing adjusts the weights of individual classes to
punish a particular class's misclassification more than the misclassification of other ones. We
disabled all three steps altogether to gain complete control over the experiments. We apply the
same data preprocessing steps to all datasets and do not perform any feature preprocessing.

Auto-Sklearn uses a list of 15 classification algorithms in its search. The list can be found on
their GitHub. We have excluded Multi-Layer Perceptron from the list as it consumes a significant
amount of resources during training, and the tabular datasets used in the experiment do not require
the use of neural networks~\cite{nn-for-tabular-data}.

\subsection{Performance Measures}

We now present a list of performance measures we used in the benchmark. We did not use metrics such
as accuracy and balanced accuracy as they are not suitable measures of the classifier's performance
in imbalanced problems~\cite{brabec2018}. In addition to the measures reported below, we also
measured F-scores and Matthews Correlation Coefficients (MCC)~\cite{matthews1975}. We do not
include them in the results due to space limitations and because F-score and MCC evaluate the
classifier only at a single operating point as opposed to the measures described below.

\subsubsection{Area Under PR Curve (PR AUC)}

Precision-Recall curve plots Precision and Recall values over possible decision thresholds. $Recall
= \frac{TP}{TP + FN}$ is plotted on the horizontal axis against $Precision = \frac{TP}{TP + FP}$ on
the vertical axis. A classifier's performance can be assessed by computing the area under the curve
(PR AUC). The PR curve does not consider the performance on the negative class and is affected by
the dataset imbalance. If the dataset imbalance differs from the real problem imbalance, the values
of PR AUC will not only be different, but the ordering of the classifiers can also
change~\cite{brabec2020}.

\subsubsection{Area Under ROC Curve (ROC AUC)}
\label{subsubsection:roc-curve}

Receiver Operating Characteristic curve plots False Positive Rate, $FPR = \frac{FP}{FP + TN}$, on
the horizontal axis against True Positive Rate, $TPR = \frac{TP}{TP + FN}$, on the vertical axis,
computed over possible decision thresholds. Again, a single number giving the performance of a
classifier can be obtained by computing the area under the curve. As FPR and TPR go against each
other, the perfect classification performance is obtained at the point $(0, 1)$. The ROC curve is
not affected by class imbalance. However, in the presence of extreme class imbalance, even a
classifier achieving high values of ROC AUC may not be practically useful~\cite{brabec2018}. The
following measure addresses this issue.

\subsubsection{Area Under Partial ROC Curve (P-ROC AUC)}

In severely imbalanced problems, even the FPR of $0.1$ may be too high for practical purposes.
Standard ROC AUC is thus affected by a large region that is irrelevant in
practice~\cite{brabec2018}. We define the \emph{Partial ROC curve} as an ROC curve with an x-axis
limited to a narrower interval to only contain values of FPR that may be practically useful.  We
then compute the area under the partial ROC curve (P-ROC AUC).

The downside is that we need to select the FPR range, and the choice is arbitrary. We decided to
follow a rule of thumb that for a practically useful classifier, the FPR should be at most as high
as the dataset's imbalance. Thus in this study, we limit the partial ROC curve's FPR range to the
interval $(0, \mathrm{dataset\_imbalance})$.

\subsection{Limitations}
\label{subsection:limitations}

The main limitation we have faced is the slowness of Python-based implementations of AutoSklearn
and preprocessing algorithms from Imbalanced Learn. Although we have previously said we did not
limit the time for the preprocessing step in any way, we were forced to abort the execution of
specific runs due to extremely long computation time. Some combinations of a preprocessing method
and a dataset ran for more than two days until we aborted them. Thus not all preprocessing methods
finished successfully on all 23 datasets. We also needed to subsample and perform feature selection
on the Ember dataset in order to complete the computation in a reasonable time and cost. The 300
most important features were chosen using a Random Forest classifier with the Gini impurity index
to rank their importance.

Another limitation we had to overcome is that publicly available cybersecurity datasets we have
found contain almost no imbalance or an imbalance that is nowhere near the real imbalance in the
domain. We have decided to randomly subsample cybersecurity datasets so that we can get closer to
the actual distribution. This decision also significantly helped us with the first problem in the
case of larger cybersecurity datasets, such as CIC-IDS-2017 and Ember. Still, even after
subsampling, the imbalance ratios are several orders of magnitude below those encountered in
practical cybersecurity systems. The datasets, however, do not contain enough data to subsample
further to achieve more realistic imbalances.

\section{Results}
\label{section:results}

\begin{figure*}
    \centering
    \includegraphics[width=0.89\linewidth]{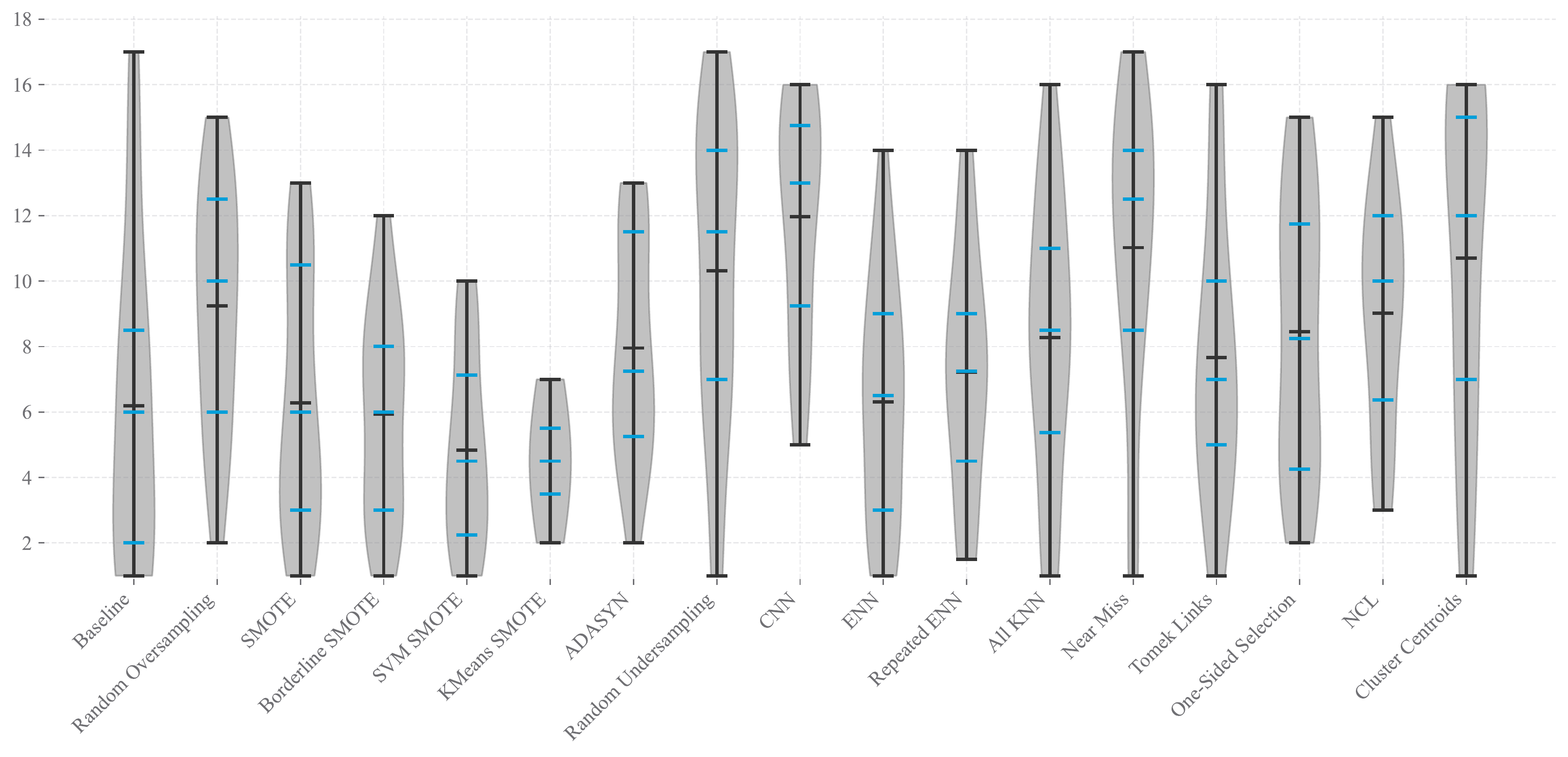}
    \caption{
        \textbf{Area under PR Curve (PR AUC).} Ranks for each method were measured across all
        datasets in the benchmark.
    }
    \label{figure:pr-auc}
\end{figure*}

\begin{figure*}
    \centering
    \includegraphics[width=0.89\linewidth]{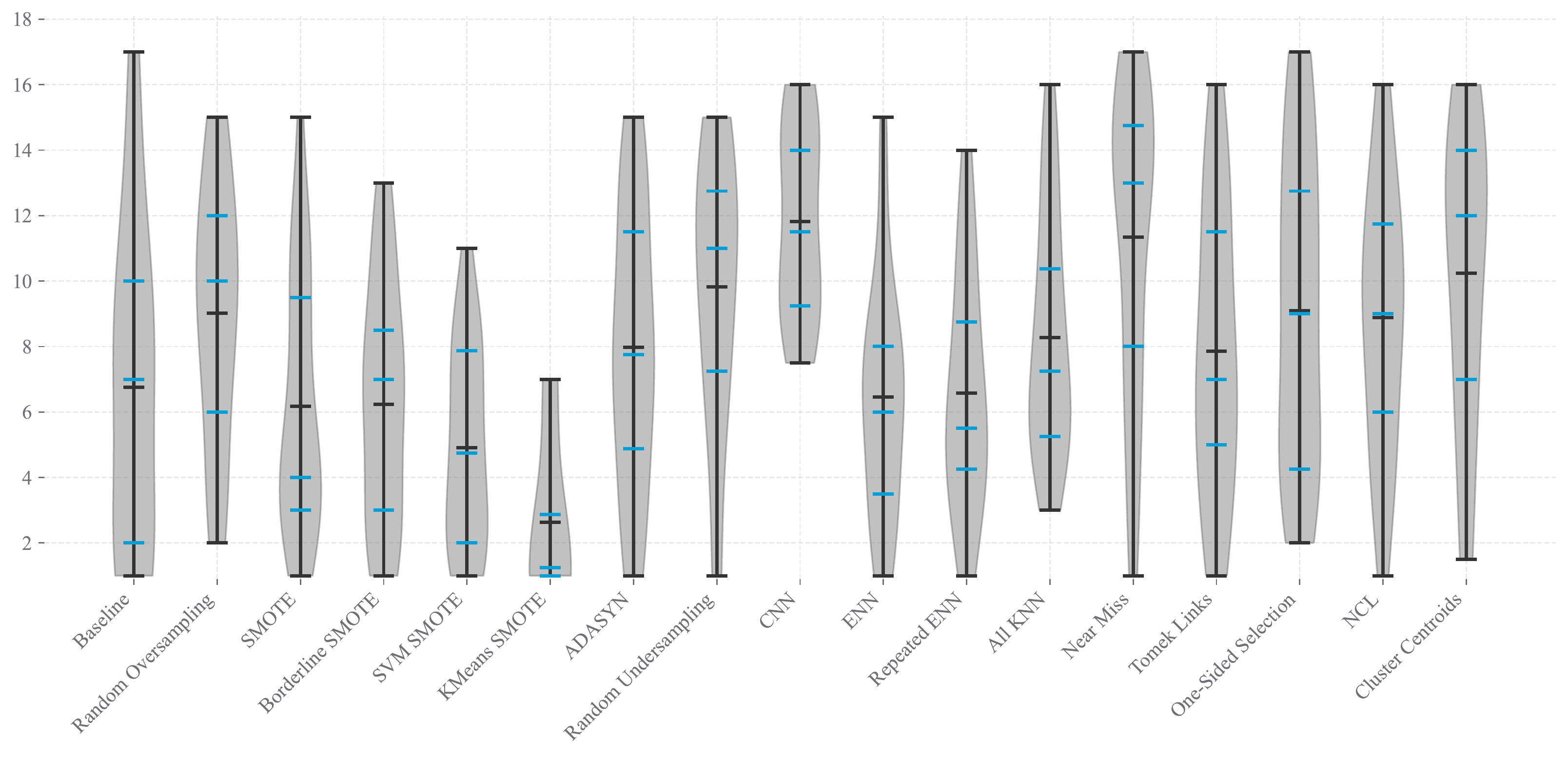}
    \caption{
        \textbf{Area under ROC Curve (ROC AUC).} Ranks for each method were measured across all
        datasets in the benchmark.
    }
    \label{figure:roc-auc}
\end{figure*}

\begin{figure*}
    \centering
    \includegraphics[width=0.89\linewidth]{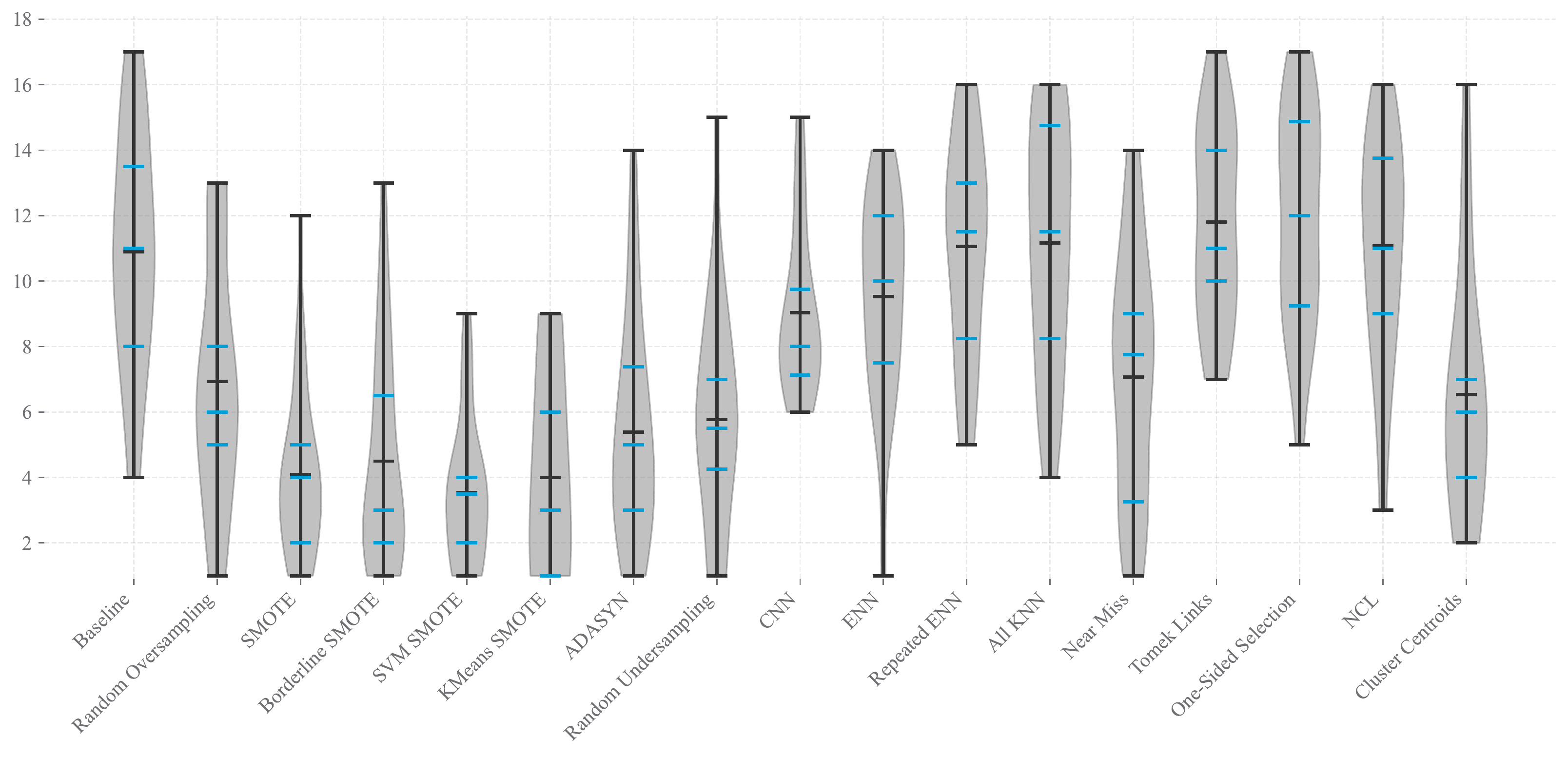}
    \caption{
        \textbf{Area under Partial ROC Curve (P-ROC AUC).} Ranks for each method were measured
        across all datasets in the benchmark.
    }
    \label{figure:partial-roc-auc}
\end{figure*}

\begin{figure*}
    \centering
    \includegraphics[width=0.89\linewidth]{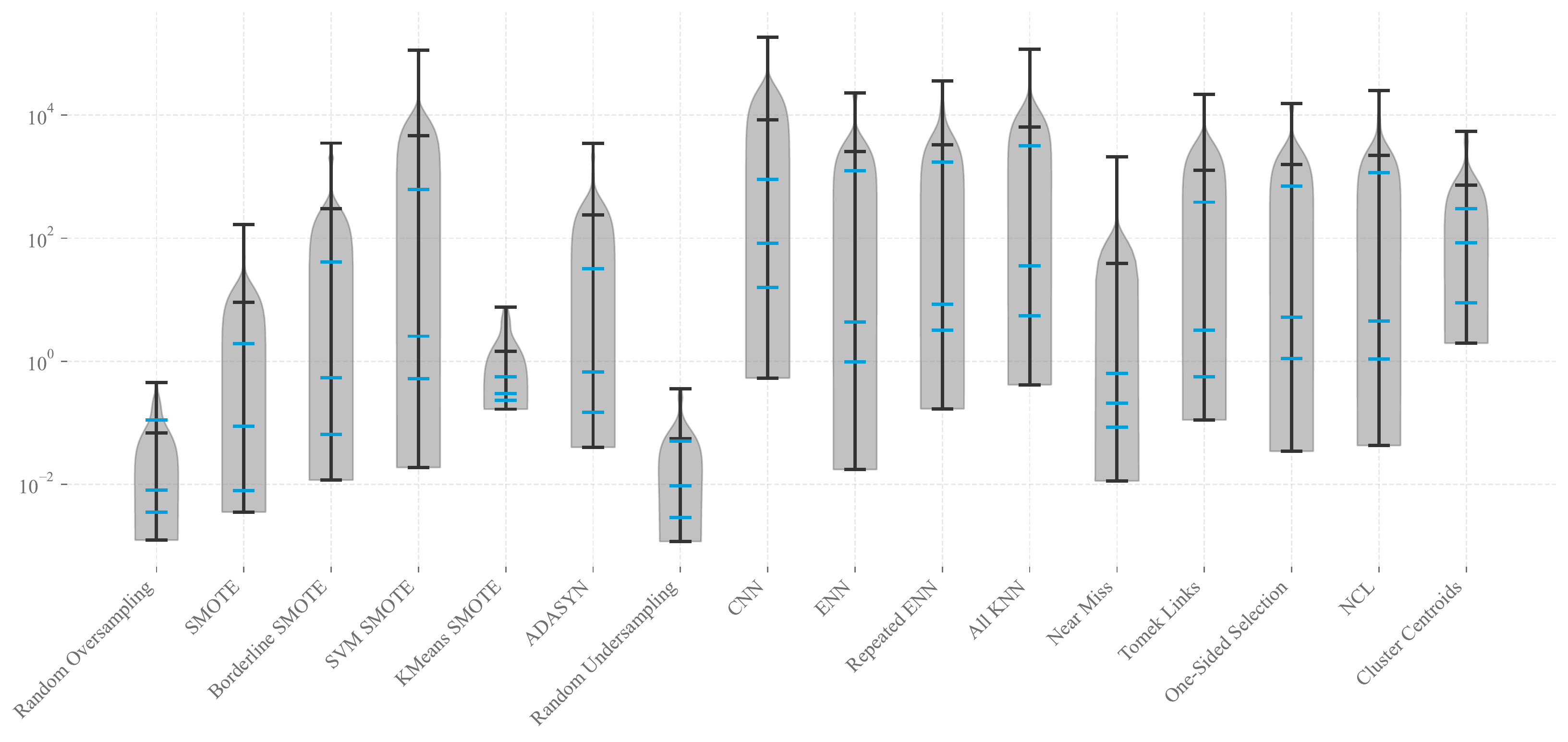}
    \caption{
        Runtimes (in seconds) for each preprocessing method were measured across all datasets in
        the benchmark.
    }
    \label{figure:preprocessing-times}
\end{figure*}

Figures~\ref{figure:pr-auc},~\ref{figure:roc-auc}~and~\ref{figure:partial-roc-auc} display the
distribution of ranks computed for each preprocessing method across all datasets in the benchmark.
Ranks were computed from metrics for each preprocessing method separately. Dark marks denote each
method's minimum, maximum and mean rank, and three blue marks indicate each method's 25th, 50th and
75th percentiles. Mean ranks are also available in Table~\ref{table:mean-rank}.
Table~\ref{table:mean-rank-cybersec} shows mean ranks computed only across the cybersecurity
datasets. We do not show distributions for cybersecurity datasets due to space limitations and a
smaller sample size. Table~\ref{table:relative-increments} zooms in on the SMOTE algorithm and its
variants and compares relative differences between these methods. Due to the small number of
successful runs, we omitted KMeansSMOTE in Table~\ref{table:relative-increments}. Lastly,
Figure~\ref{figure:preprocessing-times} shows the runtime performance of each method.

\newcolumntype{R}{>{\raggedleft\arraybackslash}X}

\begin{table}
    \centering
    \setlength\tabcolsep{2pt}

    \begin{tabularx}{\linewidth}{lRRR}
        & PR AUC & ROC AUC & P-ROC AUC \\
        \midrule
        Baseline & 6.196 // 23 & 6.761 // 23 & 10.891 // 23 \\
        Random Oversampling & 9.238 // 21 & 9.024 // 21 & 6.929 // 21 \\
        SMOTE & 6.283 // 23 & 6.174 // 23 & 4.087 // 23 \\
        Borderline SMOTE & 5.935 // 23 & 6.239 // 23 & 4.500 // 23 \\
        SVM SMOTE & 4.841 // 22 & 4.909 // 22 & 3.545 // 22 \\
        KMeans SMOTE & 4.500 // 04 & 2.625 // 04 & 4.000 // 04 \\
        ADASYN & 7.955 // 22 & 7.977 // 22 & 5.386 // 22 \\
        Random Undersampling & 10.318 // 22 & 9.818 // 22 & 5.773 // 22 \\
        CNN & 11.964 // 14 & 11.821 // 14 & 9.036 // 14 \\
        ENN & 6.310 // 21 & 6.452 // 21 & 9.524 // 21 \\
        Repeated ENN & 7.222 // 18 & 6.583 // 18 & 11.056 // 18 \\
        All KNN & 8.273 // 22 & 8.273 // 22 & 11.159 // 22 \\
        Near Miss & 11.023 // 22 & 11.341 // 22 & 7.068 // 22 \\
        Tomek Links & 7.667 // 21 & 7.857 // 21 & 11.810 // 21 \\
        One-Sided Selection & 8.455 // 22 & 9.091 // 22 & 12.000 // 22 \\
        NCL & 9.023 // 22 & 8.886 // 22 & 11.068 // 22 \\
        Cluster Centroids & 10.706 // 17 & 10.235 // 17 & 6.529 // 17 \\
    \end{tabularx}

    \vspace{4mm}

    \caption{
        \textbf{Mean Rank Across All Datasets.} The table contains average ranks for each
        combination of preprocessing method and evaluation metric computed across all datasets. The
        second number after // indicates the number of datasets used to compute the average.
    }
    \label{table:mean-rank}
\end{table}

\begin{table}
    \centering
    \setlength\tabcolsep{2pt}

    \begin{tabularx}{\linewidth}{lRRR}
        & PR AUC & ROC AUC & P-ROC AUC \\
        \midrule
        Baseline & 5.167 // 06 & 6.500 // 06 & 11.000 // 06 \\
        Random Oversampling & 8.667 // 06 & 8.667 // 06 & 7.333 // 06 \\
        SMOTE & 5.500 // 06 & 5.833 // 06 & 4.667 // 06 \\
        Borderline SMOTE & 7.000 // 06 & 8.167 // 06 & 5.667 // 06 \\
        SVM SMOTE & 5.000 // 06 & 5.000 // 06 & 4.500 // 06 \\
        KMeans SMOTE & 3.500 // 02 & 1.250 // 02 & 1.000 // 02 \\
        ADASYN & 7.500 // 06 & 8.583 // 06 & 5.667 // 06 \\
        Random Undersampling & 11.667 // 06 & 11.167 // 06 & 6.167 // 06 \\
        CNN & 14.667 // 03 & 12.000 // 03 & 10.667 // 03 \\
        ENN & 8.000 // 06 & 6.750 // 06 & 10.667 // 06 \\
        Repeated ENN & 5.800 // 05 & 5.800 // 05 & 9.800 // 05 \\
        All KNN & 9.833 // 06 & 9.500 // 06 & 11.500 // 06 \\
        Near Miss & 12.667 // 06 & 10.333 // 06 & 6.500 // 06 \\
        Tomek Links & 7.333 // 06 & 8.833 // 06 & 12.167 // 06 \\
        One-Sided Selection & 8.167 // 06 & 9.833 // 06 & 12.000 // 06 \\
        NCL & 10.333 // 06 & 10.333 // 06 & 10.333 // 06 \\
        Cluster Centroids & 9.000 // 04 & 8.125 // 04 & 6.250 // 04 \\
    \end{tabularx}

    \vspace{4mm}

    \caption{
        \textbf{Mean Rank Across Cybersecurity Datasets.} The table contains average ranks for each
        combination of preprocessing method and evaluation metric computed across cybersecurity
        datasets. The second number after // indicates the number of datasets used to compute the
        average.
    }
    \label{table:mean-rank-cybersec}
\end{table}

\begin{table}
    \centering
    \setlength\tabcolsep{2pt}

    \begin{tabularx}{0.83\linewidth}{ll|RRRRR}
        &  & \begin{turn}{90}Border. SMOTE\end{turn} & \begin{turn}{90}SVM SMOTE\end{turn} & \begin{turn}{90}ADASYN\end{turn} \\
        Method & Metric &  &  &  \\
        \midrule
        \multirow[c]{3}{*}{SMOTE} & PR AUC & 0.007 & -0.012 & -0.018 \\
        & ROC AUC & -0.012 & -0.018 & -0.012 \\
        & P-ROC AUC & -0.003 & -0.015 & -0.006 \\
        \multirow[c]{3}{*}{Border. SMOTE} & PR AUC & - & -0.020 & -0.026 \\
        & ROC AUC & - & -0.006 & -0.000 \\
        & P-ROC AUC & - & -0.013 & -0.003 \\
        \multirow[c]{3}{*}{SVM SMOTE} & PR AUC & - & - & -0.006 \\
        & ROC AUC & - & - & 0.006 \\
        & P-ROC AUC & - & - & 0.009 \\
    \end{tabularx}

    \vspace{4mm}

    \caption{
        \textbf{Incremental differences of SMOTE variants compared to each other.} The table shows
        the gains/losses of different SMOTE variants compared to each other. The results are
        differences of means computed across maximal scores attained on each dataset. The
        differences were computed by subtracting the mean score of a method in a column from the
        mean score of a method in a row.
    }
    \label{table:relative-increments}
\end{table}

\section{Discussion}
\label{section:discussion}

In this section, we take a deeper look at the results and summarise the most important findings and
recommendations. Firstly, we analyse the summary results over all datasets. Secondly, we take a
look specifically at the results on cybersecurity datasets to see whether the findings and
recommendations differ. Lastly, we discuss the computational performance of the studied methods.

To start with, let us consider the performance of the baseline method, where no preprocessing is
applied to the training dataset. The baseline method achieved a reasonable rank among all methods
and across all the measures. In the PR AUC and ROC AUC measures displayed in
Figures~\ref{figure:pr-auc} and~\ref{figure:roc-auc}, the baseline consistently ranked in the top
half of the studied methods. In the P-ROC AUC measure in Figure~\ref{figure:partial-roc-auc}, the
baseline usually ends up in the second half of methods, but it is rarely the worst method. The
baseline's performance is slightly surprising because all the methods generally claim to bring
performance gains in these types of problems. We offer several hypotheses to explain this result.
First, we are looking at summary statistics across a variety of datasets. Some methods are not
meant to be used in every scenario but are tailored for datasets with specific properties. For
example, Near Miss~\cite{mani2003} aims to remove samples at the boundary of the majority class.
This may work if these samples are mainly present due to noise, but if they are valid samples, such
removal may significantly increase the false positive rate of the classifier. Second, we perform
hyperparameter tuning of the classification layer via AutoML, which is a much stronger baseline
than usual. Furthermore, we do not guide the hyperparameter search by accuracy, which is usual, but
by ROC AUC, which is a better-suited measure for imbalanced problems. Lastly, the performance
measures we use consider multiple operating points of the final model, and PR AUC and P-ROC AUC
specifically look at the relevant operating points only. Sometimes, methods are compared via
measures such as F-score, which consider only a single operating point, and the operating point is
chosen arbitrarily without proper tuning. Tuning the operating point may have similar effects to
oversampling or undersampling.

A major takeaway is that, in general, oversampling methods outperform undersampling methods. This
pattern is visible across all performance measures and is most evident in P-ROC AUC, which we
consider to be the most practically relevant measure. Before the experiment, our intuition was that
undersampling of the majority class is one of the least preferable ways to address class imbalance
because it provides the classifier with less information to extract. The experiment's results
support this intuition. On rare occasions, undersampling may perform well. However, unless we have
a good reason to believe that it may improve a particular dataset or we have computation power to
spare, we should prefer rebalancing the dataset via oversampling. Interestingly, in the P-ROC AUC
measure, out of all the undersampling methods, simple Random Undersampling achieves the best
overall rank. This also supports the takeaway that more specialised undersampling methods are not
general-purpose tools but are best suited for niche datasets with specific properties that the
methods seek.

Several exciting observations that concern the oversampling methods can be made. In general,
oversampling methods rank better than the baseline. They outperform the baseline most convincingly
in the P-ROC AUC measure. In other measures, Random Oversampling tends to perform worse than
baseline, whereas more sophisticated oversampling methods produce better results. Contrary to the
undersampling methods, we can also see a clear trend that more sophisticated oversampling methods
outperform simple Random Oversampling. SMOTE offers the most prominent performance gain.
BorderlineSMOTE and SVM SMOTE may provide further improvements over plain SMOTE albeit smaller. We
employed the Friedman test~\cite{stats-comparison} to test whether there is a statistically
significant difference in ranks between SMOTE, BorderlineSMOTE, SVM SMOTE and ADASYN. The results
of this test ($p = 0.09$) are not statistically significant at commonly used significance levels.

We now compare the overall ranks discussed above to those achieved specifically on cybersecurity
datasets in Table~\ref{table:mean-rank-cybersec}. The ranks are similar to mean ranks across all
datasets in Table~\ref{table:mean-rank}, and the findings above tend to hold. The general trend
between oversampling and undersampling methods is similar, and SMOTE-based methods are the top
performers. However, we must highlight an important outlier in the statistics: the EMBER dataset.
On the EMBER dataset, the top-performing method in the P-ROC AUC measure was Random Undersampling
followed by Near Miss and NCL. Overall, on EMBER, undersampling methods outperformed oversampling
methods, with SMOTE, BorderlineSMOTE and SVM SMOTE achieving ranks 12, 10 and 9, respectively. The
reasons for this may be multiple. The high number of features in the EMBER dataset may hinder the
performance of methods based on nearest neighbours due to the curse of dimensionality. It is also
possible that EMBER contains a large number of noisy features which again is a complicating factor
for methods performing the nearest neighbour search. Still, even though the Random Undersampling
performed well on the EMBER dataset in this benchmark, we would not recommend relying on it in
practical systems for the classification of PE files. As we mentioned in
Section~\ref{subsection:limitations}, in the real world, we face several orders of magnitude higher
imbalance than we could introduce into the experiment. In the presence of such imbalance, we need
to achieve extremely low false positive rates and providing the classifier with a vast amount of
benign files is critical for success.

\subsection{Computational Performance}

Even from the perspective of computational performance, oversampling methods achieved slightly
better runtimes than undersampling methods. Understandably, Random Oversampling and Random
Undersampling took the least time as they do not perform heavy computations. KMeansSMOTE appears to
be faster than any of the remaining preprocessing methods. However, we need to keep in mind that
KMeansSMOTE finished successfully only on four datasets, of which two contained only a couple
hundred samples. If we do not consider these methods, we can consider the plain SMOTE as the
fastest method in the study.

On the other side of the spectrum, we see that undersampling methods are generally more
computationally intensive. SVM SMOTE and CNN stood out from the crowd and were the slowest methods
in our experiments, peaking with a maximum resampling time of almost 48 hours.

\section{Conclusion}
\label{section:conclusion}

We have conducted a novel study of 16 preprocessing methods on 23 datasets, of which six are from
the cybersecurity domain. We studied both predictive and computational performance. To that end, we
implemented a large-scale experiment which employs AutoML to consider a wide range of classifiers
and includes a hyperparameter search to remove potential sources of bias present in past
benchmarks.

Our main findings are that using dataset preprocessing when dealing with class-imbalanced
classification is often beneficial. However, at the same time, a large portion of the methods fails
to consistently outperform the baseline solution of doing nothing. Most of the time, oversampling
methods outperform undersampling methods, but exceptions exist. Among the oversampling methods, the
traditional SMOTE algorithm achieves the most significant performance gains, while its more
sophisticated variants likely lead to improvements of only incremental nature.

When we isolated our analysis only to the cybersecurity datasets which span multiple cybersecurity
domains, we reached the same conclusions as above.

Finally, it is essential to note that the method ranking is influenced by the performance measure
chosen. We include multiple performance measures that are comprehensive and suitable in practical
classification scenarios when facing class imbalance. Even though the specifics of the rankings
vary by measure, the main takeaways mentioned above are consistent.

\bibliographystyle{plain}
\bibliography{bibliography}

\clearpage
\onecolumn
\appendices

\begin{addmargin}[0.05\textwidth]{0.05\textwidth}
    \section{Additional Figures}

    Appendix A shows the distributions of ranks for metrics that we did not include in the main
    part of the paper. The distributions of ranks were computed across all datasets in the
    benchmark. Ranks were computed under various metrics for each preprocessing method separately.
    The black mark denotes each method’s mean rank, and three blue marks indicate the 25th, 50th
    and 75th percentile.

    Appendix B shows tables with the detailed results obtained on each dataset. Each table contains
    the three primary and five additional metrics - Balanced Accuracy, Precision, Recall, F1 Max
    and Matthews Correlation Coefficient. F1 Max was computed as the maximum achieved over a set of
    decision thresholds. The tables are listed in the same order as shown in
    Table~\ref{table:datasets}.
\end{addmargin}

\begin{figure}[H]
    \centering
    \includegraphics[width=0.9\linewidth]{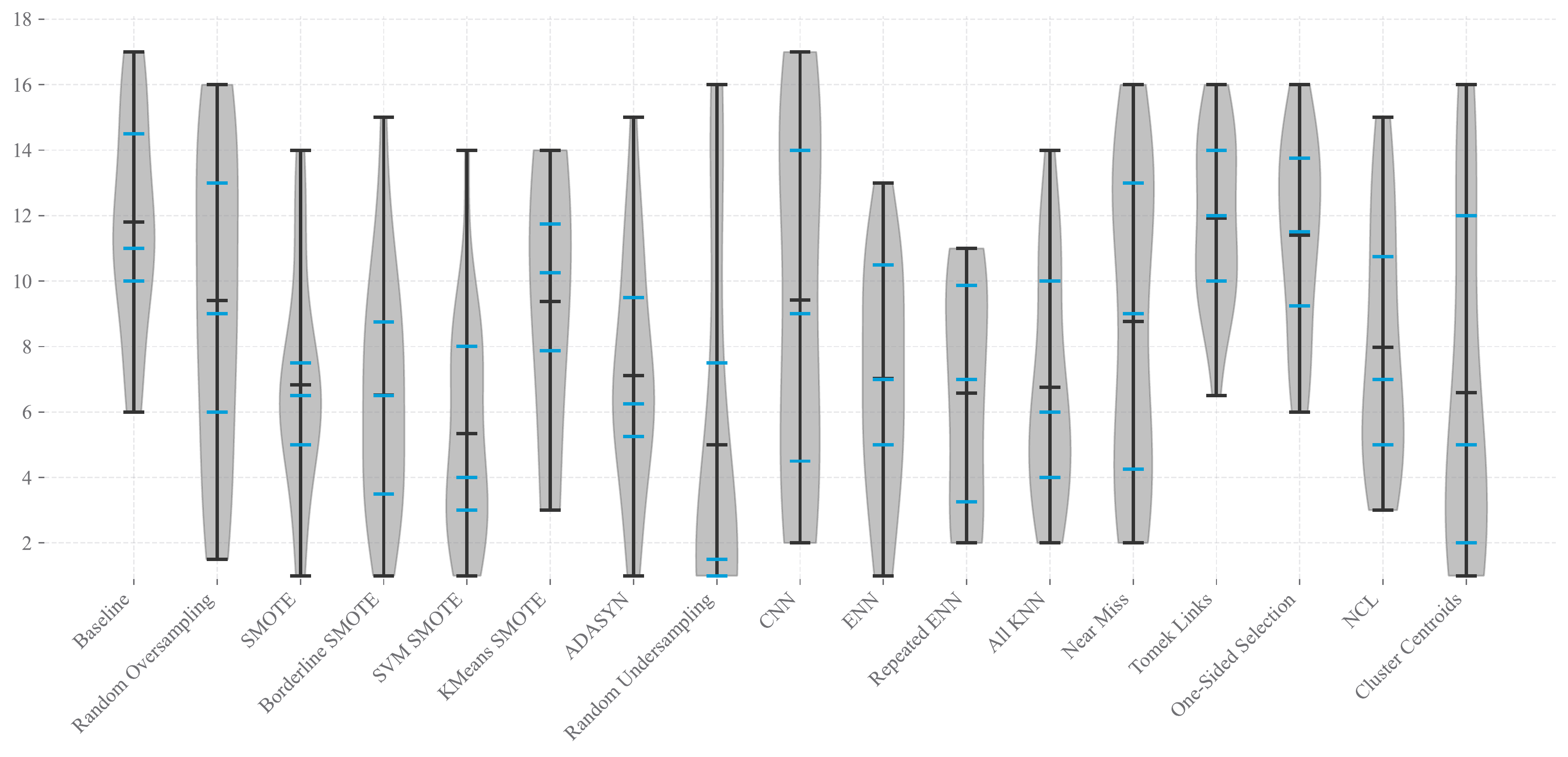}
    \caption{\textbf{Distribution of Ranks for the Balanced Accuracy Evaluation Metric.}}
\end{figure}

\begin{figure}[H]
    \centering
    \includegraphics[width=0.9\linewidth]{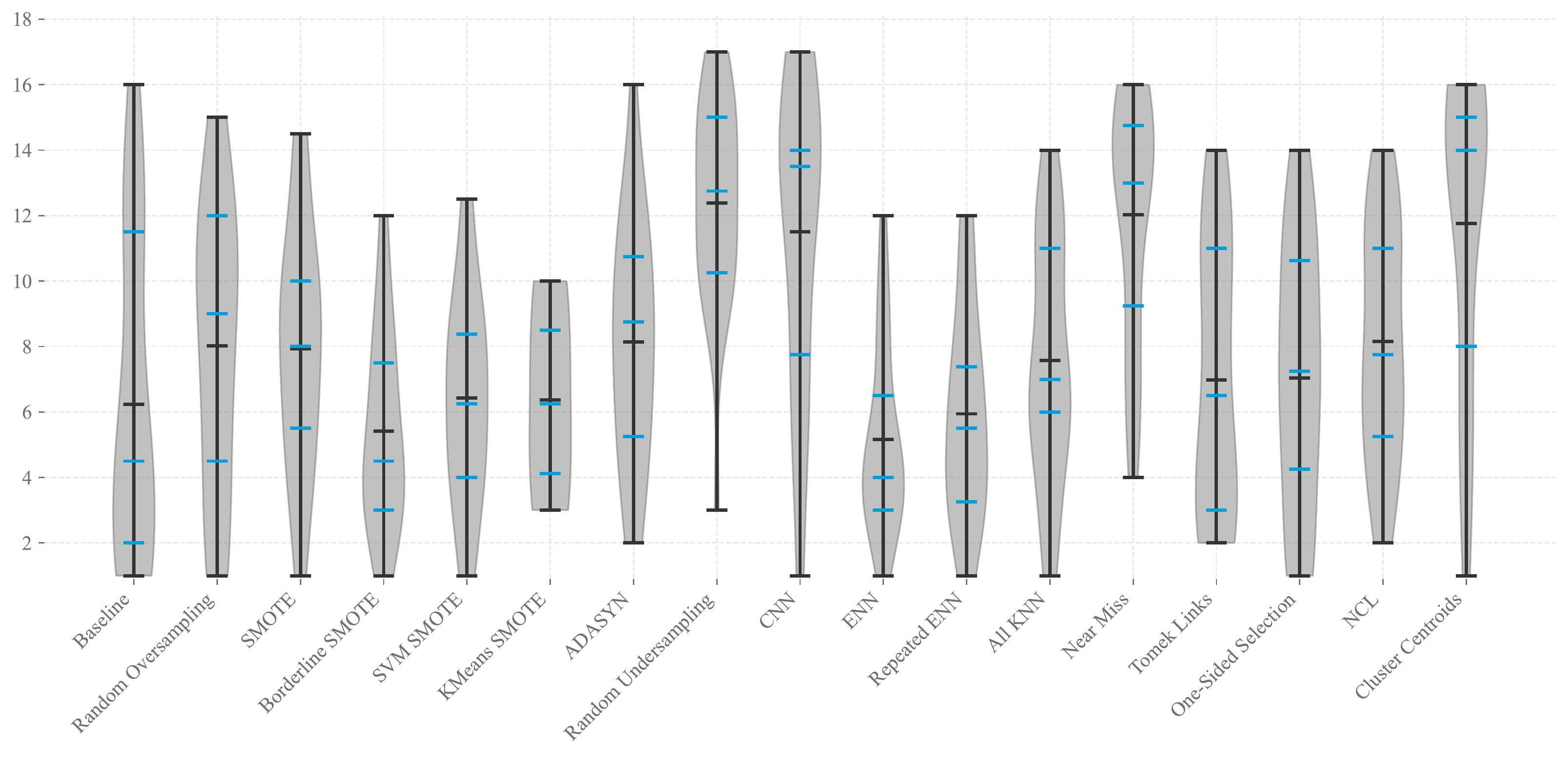}
    \caption{\textbf{Distribution of Ranks for the Precision Evaluation Metric.}}
\end{figure}

\clearpage
\begin{figure}
    \centering
    \includegraphics[width=0.9\linewidth]{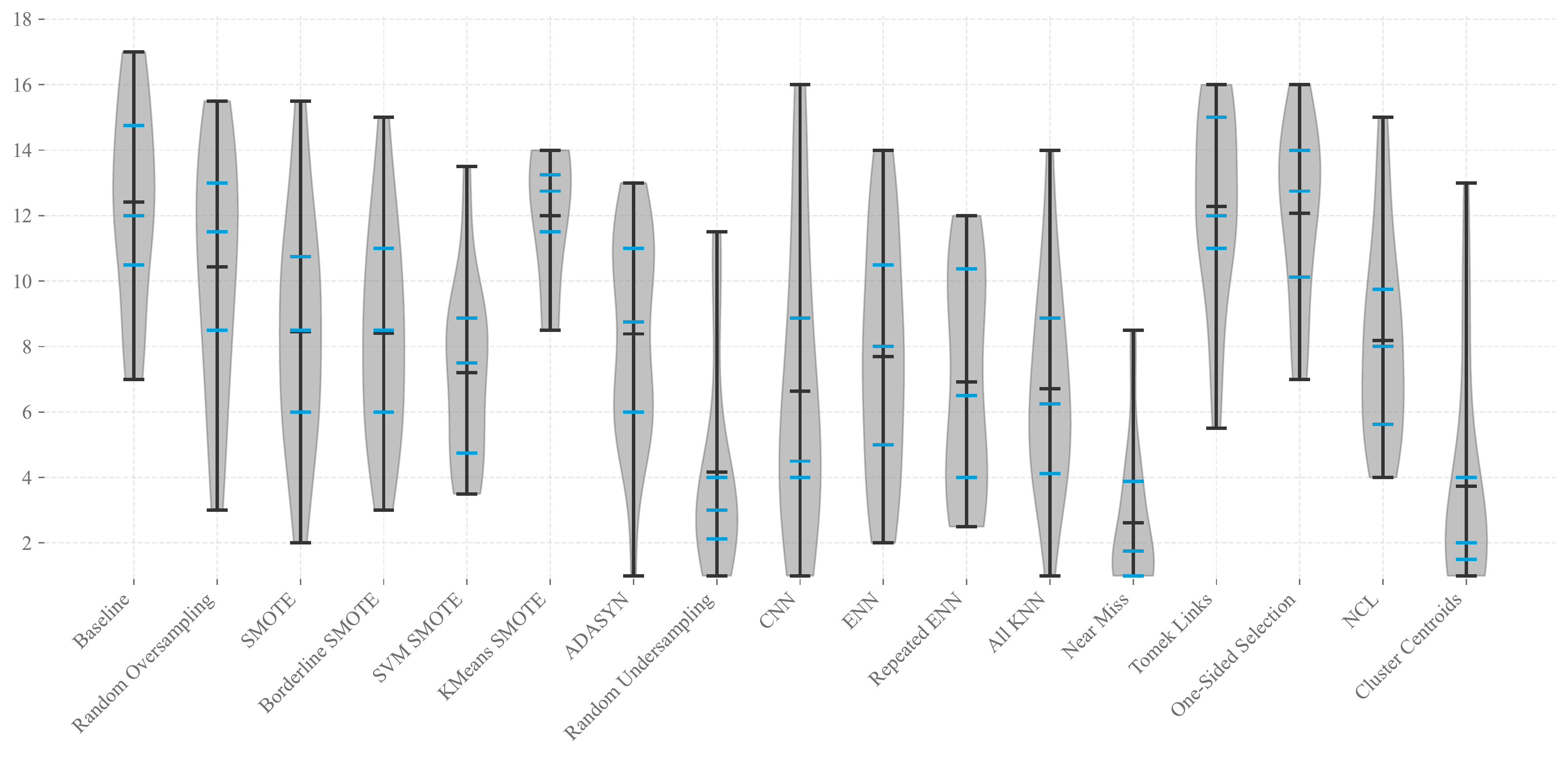}
    \caption{\textbf{Distribution of Ranks for the Recall Evaluation Metric.}}
\end{figure}

\begin{figure}
    \centering
    \includegraphics[width=0.9\linewidth]{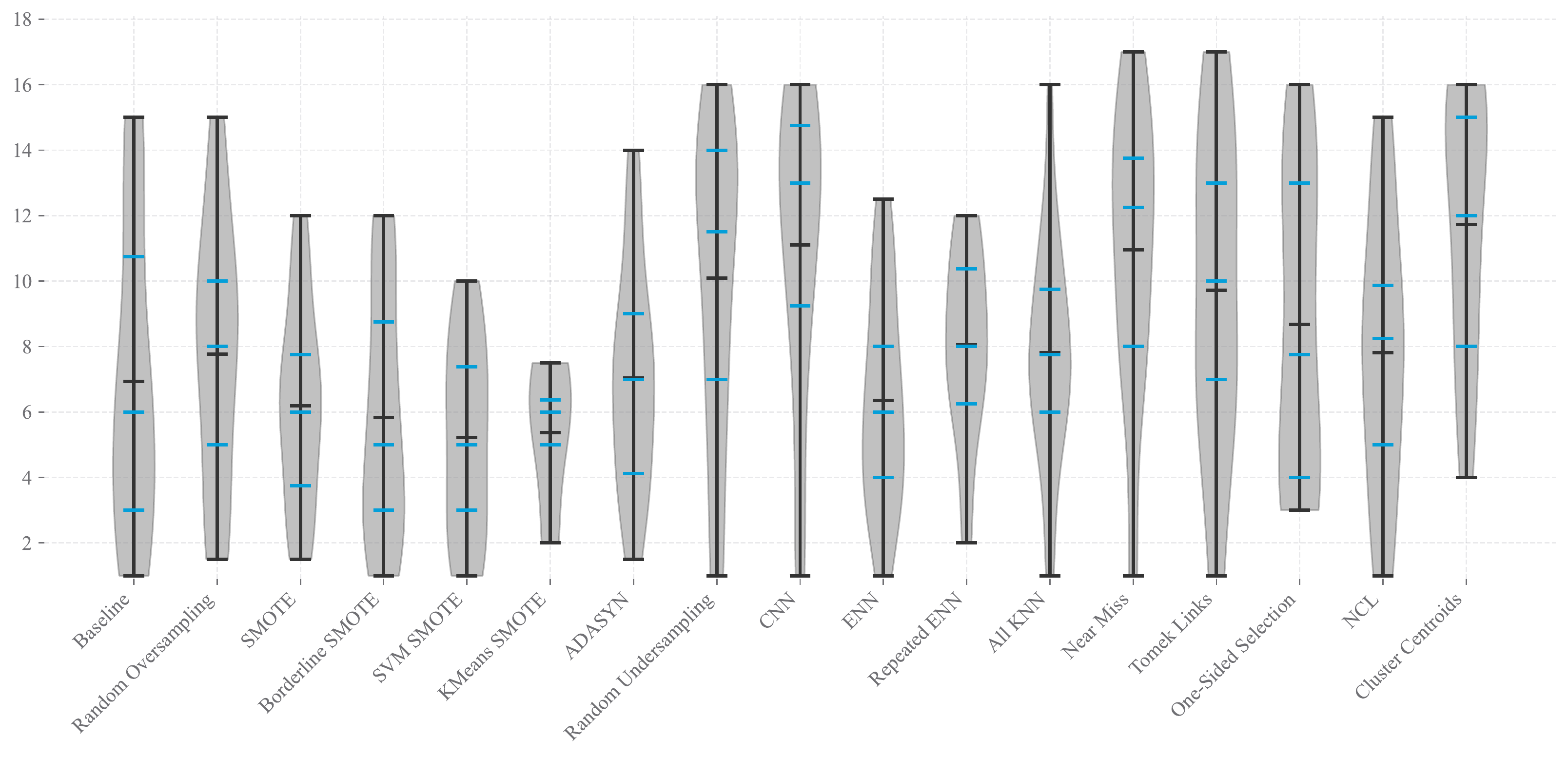}
    \caption{\textbf{Distribution of Ranks for the F1 Max Evaluation Metric.}}
\end{figure}

\clearpage
\begin{figure}[H]
    \centering
    \includegraphics[width=0.9\linewidth]{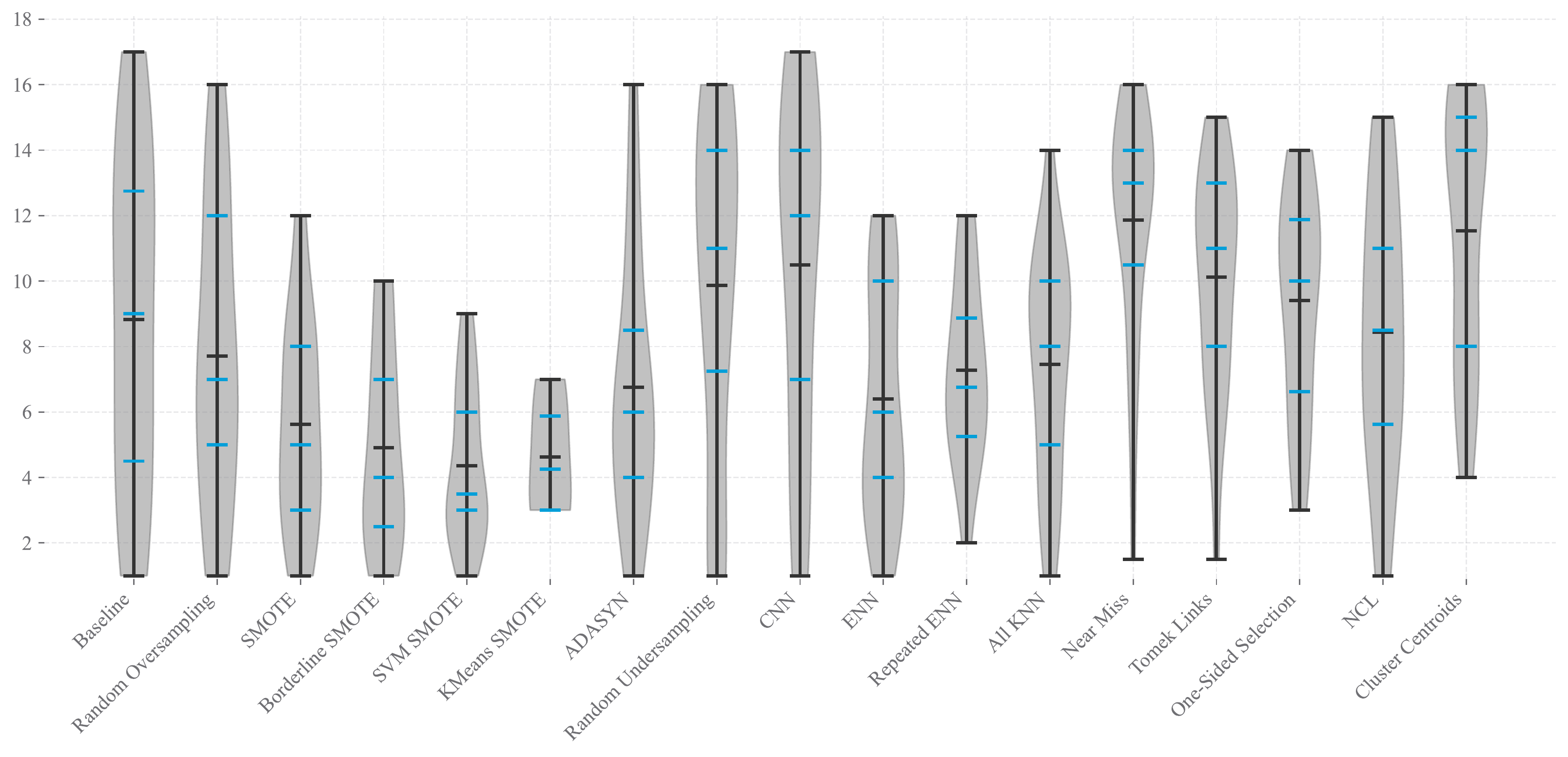}
    \caption{\textbf{Distribution of Ranks for the Matthews Correlation Coefficient Evaluation Metric.}}
\end{figure}

\clearpage
\section{Detailed Results}

\begin{table}[H]
    \centering
    \setlength\tabcolsep{2pt}
    \begin{tabularx}{\textwidth}{lRRRRRRRR}
        & B. Accuracy & Precision & Recall & F1 Max & MCC & PR AUC & ROC AUC & P-ROC AUC \\
        \midrule
        Baseline & \textbf{1.000} & \textbf{1.000} & \textbf{1.000} & \textbf{1.000} & \textbf{1.000} & \textbf{1.000} & \textbf{1.000} & \textbf{1.000} \\
        Random Oversampling & \textbf{1.000} & \textbf{1.000} & \textbf{1.000} & \textbf{1.000} & \textbf{1.000} & \textbf{1.000} & \textbf{1.000} & \textbf{1.000} \\
        SMOTE & \textbf{1.000} & \textbf{1.000} & \textbf{1.000} & \textbf{1.000} & \textbf{1.000} & \textbf{1.000} & \textbf{1.000} & \textbf{1.000} \\
        Borderline SMOTE & \textbf{1.000} & \textbf{1.000} & \textbf{1.000} & \textbf{1.000} & \textbf{1.000} & \textbf{1.000} & \textbf{1.000} & \textbf{1.000} \\
        SVM SMOTE & \textbf{1.000} & \textbf{1.000} & \textbf{1.000} & \textbf{1.000} & \textbf{1.000} & \textbf{1.000} & \textbf{1.000} & \textbf{1.000} \\
        KMeans SMOTE & N/A & N/A & N/A & N/A & N/A & N/A & N/A & N/A \\
        ADASYN & \textbf{1.000} & \textbf{1.000} & \textbf{1.000} & \textbf{1.000} & \textbf{1.000} & \textbf{1.000} & \textbf{1.000} & \textbf{1.000} \\
        Random Undersampling & 0.999 & 0.305 & \textbf{1.000} & 0.595 & 0.551 & 0.457 & 0.999 & 0.999 \\
        CNN & 1.000 & 0.696 & \textbf{1.000} & \textbf{1.000} & 0.834 & \textbf{1.000} & \textbf{1.000} & \textbf{1.000} \\
        ENN & \textbf{1.000} & \textbf{1.000} & \textbf{1.000} & \textbf{1.000} & \textbf{1.000} & \textbf{1.000} & \textbf{1.000} & \textbf{1.000} \\
        Repeated ENN & \textbf{1.000} & \textbf{1.000} & \textbf{1.000} & \textbf{1.000} & \textbf{1.000} & \textbf{1.000} & \textbf{1.000} & \textbf{1.000} \\
        All KNN & \textbf{1.000} & \textbf{1.000} & \textbf{1.000} & \textbf{1.000} & \textbf{1.000} & \textbf{1.000} & \textbf{1.000} & \textbf{1.000} \\
        Near Miss & 1.000 & 0.929 & \textbf{1.000} & \textbf{1.000} & 0.964 & \textbf{1.000} & \textbf{1.000} & \textbf{1.000} \\
        Tomek Links & \textbf{1.000} & \textbf{1.000} & \textbf{1.000} & \textbf{1.000} & \textbf{1.000} & \textbf{1.000} & \textbf{1.000} & \textbf{1.000} \\
        One-Sided Selection & \textbf{1.000} & \textbf{1.000} & \textbf{1.000} & \textbf{1.000} & \textbf{1.000} & \textbf{1.000} & \textbf{1.000} & \textbf{1.000} \\
        NCL & \textbf{1.000} & \textbf{1.000} & \textbf{1.000} & \textbf{1.000} & \textbf{1.000} & \textbf{1.000} & \textbf{1.000} & \textbf{1.000} \\
        Cluster Centroids & 0.998 & 0.291 & \textbf{1.000} & 0.582 & 0.539 & 0.436 & 0.999 & 0.999 \\
    \end{tabularx}
    \vspace{1mm}
    \caption{\textbf{Dataset Asteroid.}}
\end{table}

\begin{table}[H]
    \centering
    \setlength\tabcolsep{2pt}
    \begin{tabularx}{\textwidth}{lRRRRRRRR}
        & B. Accuracy & Precision & Recall & F1 Max & MCC & PR AUC & ROC AUC & P-ROC AUC \\
        \midrule
        Baseline & 0.500 & 0.000 & 0.000 & 0.769 & 0.000 & 0.753 & 0.916 & 0.872 \\
        Random Oversampling & \textbf{0.917} & \textbf{1.000} & 0.833 & \textbf{0.909} & \textbf{0.913} & 0.834 & 0.908 & 0.908 \\
        SMOTE & \textbf{0.917} & \textbf{1.000} & 0.833 & \textbf{0.909} & \textbf{0.913} & 0.834 & 0.932 & 0.928 \\
        Borderline SMOTE & \textbf{0.917} & \textbf{1.000} & 0.833 & \textbf{0.909} & \textbf{0.913} & 0.834 & 0.920 & 0.920 \\
        SVM SMOTE & \textbf{0.917} & \textbf{1.000} & 0.833 & \textbf{0.909} & \textbf{0.913} & \textbf{0.835} & \textbf{0.969} & \textbf{0.969} \\
        KMeans SMOTE & N/A & N/A & N/A & N/A & N/A & N/A & N/A & N/A \\
        ADASYN & \textbf{0.917} & \textbf{1.000} & 0.833 & \textbf{0.909} & \textbf{0.913} & 0.834 & 0.891 & 0.891 \\
        Random Undersampling & 0.907 & 0.066 & 0.833 & \textbf{0.909} & 0.231 & 0.834 & 0.922 & 0.922 \\
        CNN & 0.500 & 0.000 & 0.000 & 0.625 & 0.000 & 0.612 & 0.915 & 0.908 \\
        ENN & 0.833 & \textbf{1.000} & 0.667 & \textbf{0.909} & 0.816 & 0.834 & 0.868 & 0.917 \\
        Repeated ENN & 0.833 & \textbf{1.000} & 0.667 & \textbf{0.909} & 0.816 & 0.834 & 0.903 & 0.917 \\
        All KNN & 0.500 & 0.000 & 0.000 & \textbf{0.909} & 0.000 & 0.834 & 0.903 & 0.917 \\
        Near Miss & 0.916 & 0.385 & \textbf{1.000} & \textbf{0.909} & 0.565 & 0.834 & 0.957 & 0.957 \\
        Tomek Links & 0.902 & 0.045 & 0.833 & 0.204 & 0.189 & 0.097 & 0.880 & 0.531 \\
        One-Sided Selection & 0.500 & 0.000 & 0.000 & \textbf{0.909} & 0.000 & 0.834 & 0.843 & 0.917 \\
        NCL & 0.500 & 0.000 & 0.000 & \textbf{0.909} & 0.000 & 0.834 & 0.900 & 0.917 \\
        Cluster Centroids & 0.916 & 0.556 & 0.833 & \textbf{0.909} & 0.680 & 0.835 & 0.963 & 0.963 \\
    \end{tabularx}
    \vspace{1mm}
    \caption{\textbf{Dataset Credit Card Subset.}}
\end{table}

\begin{table}[H]
    \centering
    \setlength\tabcolsep{2pt}
    \begin{tabularx}{\textwidth}{lRRRRRRRR}
        & B. Accuracy & Precision & Recall & F1 Max & MCC & PR AUC & ROC AUC & P-ROC AUC \\
        \midrule
        Baseline & 0.942 & 0.055 & 0.911 & 0.276 & 0.221 & 0.149 & 0.982 & 0.549 \\
        Random Oversampling & 0.939 & \textbf{0.923} & 0.878 & \textbf{0.908} & \textbf{0.900} & 0.893 & 0.989 & 0.989 \\
        SMOTE & 0.947 & 0.827 & 0.894 & 0.902 & 0.860 & 0.897 & \textbf{0.990} & \textbf{0.990} \\
        Borderline SMOTE & 0.951 & 0.879 & 0.902 & 0.893 & 0.882 & 0.896 & 0.986 & 0.986 \\
        SVM SMOTE & 0.951 & 0.894 & 0.902 & \textbf{0.908} & 0.894 & \textbf{0.907} & 0.988 & 0.988 \\
        KMeans SMOTE & N/A & N/A & N/A & N/A & N/A & N/A & N/A & N/A \\
        ADASYN & 0.951 & 0.822 & 0.902 & 0.893 & 0.861 & 0.889 & 0.990 & 0.990 \\
        Random Undersampling & 0.951 & 0.553 & 0.919 & 0.790 & 0.702 & 0.775 & 0.987 & 0.987 \\
        CNN & 0.914 & 0.043 & 0.862 & 0.651 & 0.189 & 0.651 & 0.965 & 0.964 \\
        ENN & 0.942 & 0.884 & 0.911 & 0.882 & 0.877 & 0.868 & 0.988 & 0.940 \\
        Repeated ENN & 0.942 & 0.870 & 0.911 & 0.876 & 0.870 & 0.875 & 0.988 & 0.940 \\
        All KNN & 0.942 & 0.911 & 0.911 & 0.884 & 0.825 & 0.857 & 0.986 & 0.933 \\
        Near Miss & 0.922 & 0.026 & \textbf{0.992} & 0.863 & 0.145 & 0.844 & 0.969 & 0.967 \\
        Tomek Links & 0.942 & 0.055 & 0.911 & 0.276 & 0.221 & 0.149 & 0.982 & 0.549 \\
        One-Sided Selection & 0.942 & 0.055 & 0.911 & 0.276 & 0.221 & 0.149 & 0.982 & 0.549 \\
        NCL & 0.942 & 0.902 & 0.911 & 0.884 & 0.867 & 0.849 & 0.982 & 0.932 \\
        Cluster Centroids & \textbf{0.953} & 0.052 & 0.984 & 0.861 & 0.216 & 0.835 & 0.987 & 0.986 \\
    \end{tabularx}
    \vspace{1mm}
    \caption{\textbf{Dataset Credit Card.}}
\end{table}

\begin{table}[H]
    \centering
    \setlength\tabcolsep{2pt}
    \begin{tabularx}{\textwidth}{lRRRRRRRR}
        & B. Accuracy & Precision & Recall & F1 Max & MCC & PR AUC & ROC AUC & P-ROC AUC \\
        \midrule
        Baseline & 0.500 & 0.000 & 0.000 & 0.167 & 0.000 & 0.050 & 0.773 & 0.509 \\
        Random Oversampling & 0.491 & 0.000 & 0.000 & 0.071 & -0.009 & 0.023 & 0.877 & 0.869 \\
        SMOTE & 0.541 & 0.008 & 0.167 & 0.167 & 0.019 & 0.037 & 0.628 & 0.653 \\
        Borderline SMOTE & 0.582 & 0.167 & 0.167 & \textbf{0.286} & 0.163 & 0.097 & \textbf{0.933} & \textbf{0.929} \\
        SVM SMOTE & 0.582 & \textbf{0.200} & 0.167 & 0.267 & \textbf{0.179} & 0.085 & 0.918 & 0.908 \\
        KMeans SMOTE & N/A & N/A & N/A & N/A & N/A & N/A & N/A & N/A \\
        ADASYN & 0.498 & 0.000 & 0.000 & 0.167 & -0.004 & 0.034 & 0.705 & 0.685 \\
        Random Undersampling & \textbf{0.802} & 0.015 & 0.833 & 0.111 & 0.094 & 0.030 & 0.854 & 0.847 \\
        CNN & 0.549 & 0.005 & 0.500 & 0.048 & 0.013 & 0.007 & 0.551 & 0.513 \\
        ENN & 0.500 & 0.000 & 0.000 & 0.133 & 0.000 & 0.052 & 0.914 & 0.499 \\
        Repeated ENN & 0.500 & 0.000 & 0.000 & 0.133 & 0.000 & 0.052 & 0.914 & 0.499 \\
        All KNN & 0.500 & 0.000 & 0.000 & 0.154 & 0.000 & 0.048 & 0.911 & 0.499 \\
        Near Miss & 0.602 & 0.011 & \textbf{1.000} & 0.286 & 0.039 & \textbf{0.203} & 0.740 & 0.740 \\
        Tomek Links & 0.500 & 0.000 & 0.000 & 0.182 & 0.000 & 0.056 & 0.772 & 0.524 \\
        One-Sided Selection & 0.500 & 0.000 & 0.000 & 0.115 & 0.000 & 0.040 & 0.853 & 0.499 \\
        NCL & 0.500 & 0.000 & 0.000 & 0.143 & 0.000 & 0.042 & 0.777 & 0.499 \\
        Cluster Centroids & 0.712 & 0.009 & 0.833 & 0.222 & 0.056 & 0.075 & 0.777 & 0.787 \\
    \end{tabularx}
    \vspace{1mm}
    \caption{\textbf{Dataset PC2.}}
\end{table}

\begin{table}[H]
    \centering
    \setlength\tabcolsep{2pt}
    \begin{tabularx}{\textwidth}{lRRRRRRRR}
        & B. Accuracy & Precision & Recall & F1 Max & MCC & PR AUC & ROC AUC & P-ROC AUC \\
        \midrule
        Baseline & 0.528 & 0.167 & 0.059 & 0.389 & 0.095 & 0.246 & 0.951 & 0.634 \\
        Random Oversampling & 0.586 & 0.231 & 0.176 & 0.400 & 0.197 & 0.285 & 0.855 & 0.854 \\
        SMOTE & 0.705 & \textbf{0.583} & 0.412 & \textbf{0.581} & \textbf{0.487} & \textbf{0.501} & 0.967 & 0.964 \\
        Borderline SMOTE & 0.617 & 0.500 & 0.235 & 0.476 & 0.320 & 0.382 & \textbf{0.970} & \textbf{0.970} \\
        SVM SMOTE & 0.617 & 0.444 & 0.235 & 0.452 & 0.320 & 0.353 & 0.968 & 0.968 \\
        KMeans SMOTE & N/A & N/A & N/A & N/A & N/A & N/A & N/A & N/A \\
        ADASYN & 0.675 & 0.545 & 0.353 & \textbf{0.581} & 0.436 & 0.483 & 0.966 & 0.966 \\
        Random Undersampling & \textbf{0.838} & 0.039 & 0.824 & 0.278 & 0.158 & 0.116 & 0.927 & 0.922 \\
        CNN & 0.643 & 0.222 & 0.294 & 0.286 & 0.236 & 0.129 & 0.884 & 0.866 \\
        ENN & 0.558 & 0.500 & 0.118 & 0.437 & 0.170 & 0.250 & 0.955 & 0.641 \\
        Repeated ENN & 0.587 & 0.500 & 0.176 & 0.375 & 0.239 & 0.229 & 0.960 & 0.621 \\
        All KNN & 0.675 & 0.400 & 0.353 & 0.408 & 0.371 & 0.245 & 0.963 & 0.621 \\
        Near Miss & 0.814 & 0.039 & \textbf{1.000} & 0.192 & 0.152 & 0.132 & 0.844 & 0.844 \\
        Tomek Links & 0.528 & 0.167 & 0.059 & 0.414 & 0.095 & 0.285 & 0.958 & 0.641 \\
        One-Sided Selection & 0.558 & 0.286 & 0.118 & 0.483 & 0.180 & 0.275 & 0.928 & 0.658 \\
        NCL & 0.528 & 0.167 & 0.059 & 0.400 & 0.095 & 0.219 & 0.956 & 0.624 \\
        Cluster Centroids & 0.723 & 0.100 & \textbf{1.000} & 0.189 & 0.206 & 0.110 & 0.911 & 0.911 \\
    \end{tabularx}
    \vspace{1mm}
    \caption{\textbf{Dataset MC1.}}
\end{table}

\begin{table}[H]
    \centering
    \setlength\tabcolsep{2pt}
    \begin{tabularx}{\textwidth}{lRRRRRRRR}
        & B. Accuracy & Precision & Recall & F1 Max & MCC & PR AUC & ROC AUC & P-ROC AUC \\
        \midrule
        Baseline & 0.500 & 0.000 & 0.000 & 0.188 & -0.002 & 0.096 & 0.807 & 0.563 \\
        Random Oversampling & 0.501 & 0.019 & 0.008 & 0.157 & 0.003 & 0.059 & 0.786 & 0.782 \\
        SMOTE & 0.538 & 0.200 & 0.081 & 0.204 & 0.119 & 0.123 & 0.883 & 0.879 \\
        Borderline SMOTE & 0.547 & 0.294 & 0.097 & 0.220 & 0.162 & \textbf{0.135} & 0.845 & 0.845 \\
        SVM SMOTE & 0.547 & 0.300 & 0.097 & 0.183 & 0.164 & 0.124 & 0.864 & 0.852 \\
        KMeans SMOTE & N/A & N/A & N/A & N/A & N/A & N/A & N/A & N/A \\
        ADASYN & 0.537 & 0.176 & 0.081 & 0.205 & 0.108 & 0.117 & \textbf{0.885} & \textbf{0.880} \\
        Random Undersampling & \textbf{0.683} & 0.048 & 0.589 & 0.222 & 0.107 & 0.093 & 0.758 & 0.758 \\
        CNN & 0.605 & \textbf{0.667} & 0.218 & \textbf{0.248} & \textbf{0.233} & 0.099 & 0.714 & 0.672 \\
        ENN & 0.552 & 0.093 & 0.121 & 0.154 & 0.091 & 0.063 & 0.805 & 0.535 \\
        Repeated ENN & 0.585 & 0.082 & 0.210 & 0.154 & 0.101 & 0.086 & 0.821 & 0.549 \\
        All KNN & 0.578 & 0.134 & 0.194 & 0.158 & 0.127 & 0.074 & 0.812 & 0.548 \\
        Near Miss & 0.589 & 0.020 & \textbf{0.984} & 0.080 & 0.046 & 0.028 & 0.637 & 0.637 \\
        Tomek Links & 0.506 & 0.061 & 0.016 & 0.142 & 0.024 & 0.064 & 0.810 & 0.528 \\
        One-Sided Selection & 0.502 & 0.033 & 0.008 & 0.139 & 0.009 & 0.061 & 0.802 & 0.528 \\
        NCL & 0.563 & 0.101 & 0.145 & 0.143 & 0.106 & 0.063 & 0.801 & 0.543 \\
        Cluster Centroids & 0.605 & 0.022 & 0.976 & 0.099 & 0.052 & 0.035 & 0.690 & 0.690 \\
    \end{tabularx}
    \vspace{1mm}
    \caption{\textbf{Dataset Employee Turnover.}}
\end{table}

\begin{table}[H]
    \centering
    \setlength\tabcolsep{2pt}
    \begin{tabularx}{\textwidth}{lRRRRRRRR}
        & B. Accuracy & Precision & Recall & F1 Max & MCC & PR AUC & ROC AUC & P-ROC AUC \\
        \midrule
        Baseline & 0.553 & \textbf{1.000} & 0.105 & 0.698 & 0.322 & 0.633 & 0.956 & 0.800 \\
        Random Oversampling & 0.789 & \textbf{1.000} & 0.579 & 0.788 & 0.758 & 0.805 & 0.993 & 0.861 \\
        SMOTE & 0.868 & 0.929 & 0.737 & 0.824 & 0.800 & 0.867 & 0.995 & 0.900 \\
        Borderline SMOTE & 0.868 & \textbf{1.000} & 0.737 & \textbf{0.848} & \textbf{0.857} & 0.851 & 0.995 & 0.896 \\
        SVM SMOTE & 0.866 & \textbf{1.000} & 0.737 & 0.813 & 0.825 & 0.856 & 0.994 & 0.900 \\
        KMeans SMOTE & 0.789 & \textbf{1.000} & 0.579 & 0.813 & 0.725 & 0.865 & \textbf{0.996} & 0.897 \\
        ADASYN & 0.868 & 0.933 & 0.737 & 0.824 & 0.827 & 0.855 & 0.996 & \textbf{0.996} \\
        Random Undersampling & 0.949 & 0.237 & 0.947 & 0.710 & 0.446 & 0.766 & 0.989 & 0.988 \\
        CNN & 0.881 & 0.560 & 0.789 & 0.788 & 0.636 & 0.769 & 0.985 & 0.856 \\
        ENN & 0.841 & \textbf{1.000} & 0.684 & 0.813 & 0.723 & \textbf{0.867} & 0.995 & 0.904 \\
        Repeated ENN & 0.841 & \textbf{1.000} & 0.684 & 0.813 & 0.723 & \textbf{0.867} & 0.995 & 0.904 \\
        All KNN & 0.840 & 0.722 & 0.684 & 0.750 & 0.699 & 0.790 & 0.977 & 0.851 \\
        Near Miss & 0.889 & 0.484 & \textbf{1.000} & 0.774 & 0.611 & 0.773 & 0.979 & 0.979 \\
        Tomek Links & 0.760 & 0.588 & 0.526 & 0.621 & 0.550 & 0.688 & 0.967 & 0.788 \\
        One-Sided Selection & 0.787 & 0.647 & 0.579 & 0.629 & 0.607 & 0.695 & 0.971 & 0.798 \\
        NCL & 0.815 & \textbf{1.000} & 0.632 & 0.813 & 0.758 & 0.831 & 0.994 & 0.874 \\
        Cluster Centroids & \textbf{0.961} & 0.367 & 0.947 & 0.788 & 0.582 & 0.830 & 0.995 & 0.995 \\
    \end{tabularx}
    \vspace{1mm}
    \caption{\textbf{Dataset Satellite.}}
\end{table}

\begin{table}[H]
    \centering
    \setlength\tabcolsep{2pt}
    \begin{tabularx}{\textwidth}{lRRRRRRRR}
        & B. Accuracy & Precision & Recall & F1 Max & MCC & PR AUC & ROC AUC & P-ROC AUC \\
        \midrule
        Baseline & 0.773 & \textbf{0.778} & 0.549 & 0.662 & \textbf{0.647} & 0.717 & 0.976 & 0.833 \\
        Random Oversampling & 0.883 & 0.275 & 0.818 & 0.604 & 0.456 & 0.561 & 0.895 & 0.895 \\
        SMOTE & 0.844 & 0.479 & 0.715 & 0.591 & 0.567 & 0.634 & 0.966 & 0.965 \\
        Borderline SMOTE & 0.837 & 0.558 & 0.694 & 0.602 & 0.592 & 0.625 & 0.972 & 0.970 \\
        SVM SMOTE & 0.835 & 0.541 & 0.688 & 0.602 & 0.592 & 0.623 & 0.971 & 0.969 \\
        KMeans SMOTE & N/A & N/A & N/A & N/A & N/A & N/A & N/A & N/A \\
        ADASYN & 0.842 & 0.417 & 0.718 & 0.579 & 0.527 & 0.618 & 0.963 & 0.961 \\
        Random Undersampling & \textbf{0.927} & 0.240 & 0.930 & \textbf{0.677} & 0.449 & \textbf{0.738} & \textbf{0.979} & \textbf{0.979} \\
        CNN & N/A & N/A & N/A & N/A & N/A & N/A & N/A & N/A \\
        ENN & N/A & N/A & N/A & N/A & N/A & N/A & N/A & N/A \\
        Repeated ENN & N/A & N/A & N/A & N/A & N/A & N/A & N/A & N/A \\
        All KNN & 0.881 & 0.331 & 0.799 & 0.584 & 0.498 & 0.602 & 0.970 & 0.783 \\
        Near Miss & 0.812 & 0.107 & \textbf{0.937} & 0.327 & 0.251 & 0.251 & 0.906 & 0.900 \\
        Tomek Links & N/A & N/A & N/A & N/A & N/A & N/A & N/A & N/A \\
        One-Sided Selection & N/A & N/A & N/A & N/A & N/A & N/A & N/A & N/A \\
        NCL & 0.860 & 0.623 & 0.735 & 0.656 & 0.642 & 0.701 & 0.976 & 0.829 \\
        Cluster Centroids & N/A & N/A & N/A & N/A & N/A & N/A & N/A & N/A \\
    \end{tabularx}
    \vspace{1mm}
    \caption{\textbf{Dataset BNG - Solar Flare.}}
\end{table}

\begin{table}[H]
    \centering
    \setlength\tabcolsep{2pt}
    \begin{tabularx}{\textwidth}{lRRRRRRRR}
        & B. Accuracy & Precision & Recall & F1 Max & MCC & PR AUC & ROC AUC & P-ROC AUC \\
        \midrule
        Baseline & 0.768 & \textbf{0.875} & 0.538 & 0.723 & 0.681 & \textbf{0.790} & 0.967 & 0.867 \\
        Random Oversampling & 0.829 & 0.867 & 0.662 & 0.736 & \textbf{0.716} & 0.675 & 0.911 & 0.911 \\
        SMOTE & 0.879 & 0.633 & 0.769 & 0.716 & 0.690 & 0.769 & 0.975 & 0.973 \\
        Borderline SMOTE & 0.886 & 0.695 & 0.785 & 0.699 & 0.678 & 0.771 & 0.975 & 0.975 \\
        SVM SMOTE & 0.887 & 0.658 & 0.785 & 0.718 & 0.691 & 0.785 & \textbf{0.977} & \textbf{0.975} \\
        KMeans SMOTE & N/A & N/A & N/A & N/A & N/A & N/A & N/A & N/A \\
        ADASYN & 0.846 & 0.587 & 0.708 & 0.667 & 0.621 & 0.716 & 0.956 & 0.953 \\
        Random Undersampling & \textbf{0.935} & 0.246 & 0.938 & \textbf{0.738} & 0.461 & 0.762 & 0.958 & 0.957 \\
        CNN & 0.823 & 0.506 & 0.800 & 0.626 & 0.567 & 0.657 & 0.926 & 0.922 \\
        ENN & 0.805 & 0.837 & 0.615 & 0.719 & 0.675 & 0.775 & 0.968 & 0.865 \\
        Repeated ENN & 0.884 & 0.818 & 0.785 & 0.718 & 0.667 & 0.761 & 0.968 & 0.861 \\
        All KNN & 0.863 & 0.787 & 0.738 & 0.708 & 0.663 & 0.734 & 0.967 & 0.854 \\
        Near Miss & 0.717 & 0.047 & \textbf{0.954} & 0.571 & 0.132 & 0.550 & 0.900 & 0.900 \\
        Tomek Links & 0.768 & 0.833 & 0.538 & 0.700 & 0.664 & 0.781 & 0.968 & 0.868 \\
        One-Sided Selection & 0.760 & 0.829 & 0.523 & 0.714 & 0.653 & 0.767 & 0.967 & 0.862 \\
        NCL & 0.848 & 0.760 & 0.708 & 0.692 & 0.660 & 0.747 & 0.971 & 0.862 \\
        Cluster Centroids & 0.781 & 0.057 & \textbf{0.954} & 0.698 & 0.175 & 0.645 & 0.958 & 0.958 \\
    \end{tabularx}
    \vspace{1mm}
    \caption{\textbf{Dataset Mammography.}}
\end{table}

\begin{table}[H]
    \centering
    \setlength\tabcolsep{2pt}
    \begin{tabularx}{\textwidth}{lRRRRRRRR}
        & B. Accuracy & Precision & Recall & F1 Max & MCC & PR AUC & ROC AUC & P-ROC AUC \\
        \midrule
        Baseline & 0.995 & 0.995 & 0.990 & 0.995 & 0.992 & 1.000 & 1.000 & 1.000 \\
        Random Oversampling & 0.995 & 0.995 & 0.990 & 0.995 & 0.987 & 1.000 & 1.000 & 1.000 \\
        SMOTE & 0.997 & 0.995 & 0.995 & 0.998 & 0.995 & 1.000 & 1.000 & 1.000 \\
        Borderline SMOTE & 0.998 & \textbf{1.000} & 0.995 & 0.998 & 0.997 & 1.000 & 1.000 & 1.000 \\
        SVM SMOTE & 1.000 & 0.995 & \textbf{1.000} & 0.998 & 0.995 & 1.000 & 1.000 & 1.000 \\
        KMeans SMOTE & N/A & N/A & N/A & N/A & N/A & N/A & N/A & N/A \\
        ADASYN & 0.997 & 0.995 & 0.995 & 0.995 & 0.995 & 1.000 & 1.000 & 1.000 \\
        Random Undersampling & 0.995 & 0.822 & \textbf{1.000} & 0.990 & 0.902 & 0.999 & 1.000 & 1.000 \\
        CNN & 1.000 & 0.995 & \textbf{1.000} & 0.998 & 0.997 & 1.000 & 1.000 & 1.000 \\
        ENN & \textbf{1.000} & \textbf{1.000} & \textbf{1.000} & \textbf{1.000} & \textbf{1.000} & \textbf{1.000} & \textbf{1.000} & \textbf{1.000} \\
        Repeated ENN & 0.997 & \textbf{1.000} & 0.995 & 0.995 & 0.995 & 1.000 & 1.000 & 1.000 \\
        All KNN & 0.997 & \textbf{1.000} & 0.995 & 0.998 & 0.992 & 1.000 & 1.000 & 1.000 \\
        Near Miss & 0.999 & 0.949 & \textbf{1.000} & 0.995 & 0.973 & 1.000 & 1.000 & 1.000 \\
        Tomek Links & 0.995 & \textbf{1.000} & 0.990 & 0.995 & 0.995 & 1.000 & 1.000 & 1.000 \\
        One-Sided Selection & 0.997 & 0.995 & 0.995 & 0.995 & 0.995 & 1.000 & 1.000 & 1.000 \\
        NCL & 0.998 & 0.990 & \textbf{1.000} & 0.993 & 0.990 & 1.000 & 1.000 & 1.000 \\
        Cluster Centroids & 0.996 & 0.857 & \textbf{1.000} & 0.990 & 0.922 & 0.999 & 1.000 & 1.000 \\
    \end{tabularx}
    \vspace{1mm}
    \caption{\textbf{Dataset Letter.}}
\end{table}

\begin{table}[H]
    \centering
    \setlength\tabcolsep{2pt}
    \begin{tabularx}{\textwidth}{lRRRRRRRR}
        & B. Accuracy & Precision & Recall & F1 Max & MCC & PR AUC & ROC AUC & P-ROC AUC \\
        \midrule
        Baseline & 0.998 & 0.994 & 0.996 & 0.996 & 0.994 & 1.000 & 1.000 & 1.000 \\
        Random Oversampling & 0.999 & 0.988 & 0.998 & 0.995 & 0.992 & 1.000 & 1.000 & 1.000 \\
        SMOTE & 0.998 & 0.989 & 0.997 & \textbf{0.997} & 0.992 & 1.000 & 1.000 & 1.000 \\
        Borderline SMOTE & 0.998 & \textbf{0.996} & 0.996 & \textbf{0.997} & \textbf{0.996} & 1.000 & 1.000 & 1.000 \\
        SVM SMOTE & \textbf{0.999} & 0.991 & 0.999 & \textbf{0.997} & 0.994 & \textbf{1.000} & \textbf{1.000} & \textbf{1.000} \\
        KMeans SMOTE & N/A & N/A & N/A & N/A & N/A & N/A & N/A & N/A \\
        ADASYN & 0.998 & 0.991 & 0.997 & 0.995 & 0.994 & 1.000 & 1.000 & 1.000 \\
        Random Undersampling & 0.996 & 0.882 & 0.999 & 0.969 & 0.936 & 0.995 & 1.000 & 1.000 \\
        CNN & N/A & N/A & N/A & N/A & N/A & N/A & N/A & N/A \\
        ENN & N/A & N/A & N/A & N/A & N/A & N/A & N/A & N/A \\
        Repeated ENN & 0.999 & 0.981 & 0.999 & 0.991 & 0.989 & 0.999 & 1.000 & 0.999 \\
        All KNN & 0.999 & 0.984 & 0.998 & 0.993 & 0.990 & 1.000 & 1.000 & 1.000 \\
        Near Miss & 0.984 & 0.585 & 0.999 & 0.987 & 0.753 & 0.997 & 1.000 & 1.000 \\
        Tomek Links & 0.998 & 0.995 & 0.996 & 0.996 & \textbf{0.996} & 1.000 & 1.000 & 1.000 \\
        One-Sided Selection & 0.996 & 0.991 & 0.993 & 0.992 & 0.991 & 0.999 & 1.000 & 1.000 \\
        NCL & 0.998 & 0.984 & 0.997 & 0.993 & 0.990 & 0.999 & 1.000 & 0.999 \\
        Cluster Centroids & 0.999 & 0.989 & \textbf{1.000} & 0.996 & 0.993 & 1.000 & 1.000 & 1.000 \\
    \end{tabularx}
    \vspace{1mm}
    \caption{\textbf{Dataset Relevant Images.}}
\end{table}

\begin{table}[H]
    \centering
    \setlength\tabcolsep{2pt}
    \begin{tabularx}{\textwidth}{lRRRRRRRR}
        & B. Accuracy & Precision & Recall & F1 Max & MCC & PR AUC & ROC AUC & P-ROC AUC \\
        \midrule
        Baseline & 0.507 & 0.455 & 0.014 & 0.193 & 0.075 & 0.135 & \textbf{0.721} & 0.556 \\
        Random Oversampling & N/A & N/A & N/A & N/A & N/A & N/A & N/A & N/A \\
        SMOTE & 0.595 & 0.091 & 0.473 & 0.144 & 0.095 & 0.078 & 0.638 & 0.635 \\
        Borderline SMOTE & 0.601 & 0.102 & 0.432 & 0.147 & 0.099 & 0.087 & 0.653 & 0.651 \\
        SVM SMOTE & 0.598 & 0.102 & 0.375 & 0.148 & 0.104 & 0.094 & 0.657 & 0.656 \\
        KMeans SMOTE & N/A & N/A & N/A & N/A & N/A & N/A & N/A & N/A \\
        ADASYN & 0.584 & 0.090 & 0.316 & 0.142 & 0.095 & 0.077 & 0.634 & 0.620 \\
        Random Undersampling & \textbf{0.653} & 0.097 & \textbf{0.653} & 0.187 & \textbf{0.141} & 0.132 & 0.716 & \textbf{0.716} \\
        CNN & N/A & N/A & N/A & N/A & N/A & N/A & N/A & N/A \\
        ENN & 0.543 & \textbf{0.539} & 0.100 & 0.191 & 0.134 & 0.140 & 0.718 & 0.565 \\
        Repeated ENN & 0.608 & 0.449 & 0.342 & 0.193 & 0.131 & 0.136 & 0.720 & 0.564 \\
        All KNN & 0.595 & 0.445 & 0.272 & \textbf{0.195} & 0.136 & \textbf{0.140} & 0.719 & 0.566 \\
        Near Miss & N/A & N/A & N/A & N/A & N/A & N/A & N/A & N/A \\
        Tomek Links & 0.509 & 0.462 & 0.019 & 0.191 & 0.087 & 0.135 & 0.720 & 0.557 \\
        One-Sided Selection & 0.510 & 0.496 & 0.020 & 0.193 & 0.093 & 0.136 & 0.720 & 0.557 \\
        NCL & 0.538 & 0.359 & 0.089 & 0.192 & 0.129 & 0.138 & 0.718 & 0.564 \\
        Cluster Centroids & N/A & N/A & N/A & N/A & N/A & N/A & N/A & N/A \\
    \end{tabularx}
    \vspace{1mm}
    \caption{\textbf{Dataset Click Prediction V1.}}
\end{table}

\begin{table}[H]
    \centering
    \setlength\tabcolsep{2pt}
    \begin{tabularx}{\textwidth}{lRRRRRRRR}
        & B. Accuracy & Precision & Recall & F1 Max & MCC & PR AUC & ROC AUC & P-ROC AUC \\
        \midrule
        Baseline & 0.506 & 0.656 & 0.013 & 0.186 & 0.087 & 0.133 & 0.711 & 0.551 \\
        Random Oversampling & 0.522 & 0.184 & 0.054 & 0.147 & 0.078 & 0.095 & 0.659 & 0.659 \\
        SMOTE & 0.535 & 0.121 & 0.115 & 0.138 & 0.072 & 0.089 & 0.651 & 0.651 \\
        Borderline SMOTE & 0.541 & 0.149 & 0.112 & 0.155 & 0.094 & 0.096 & 0.670 & 0.659 \\
        SVM SMOTE & 0.539 & 0.158 & 0.104 & 0.153 & 0.095 & 0.100 & 0.667 & 0.656 \\
        KMeans SMOTE & N/A & N/A & N/A & N/A & N/A & N/A & N/A & N/A \\
        ADASYN & N/A & N/A & N/A & N/A & N/A & N/A & N/A & N/A \\
        Random Undersampling & \textbf{0.642} & 0.098 & 0.639 & 0.177 & \textbf{0.133} & 0.119 & 0.702 & \textbf{0.699} \\
        CNN & 0.532 & 0.225 & 0.075 & 0.152 & 0.107 & 0.103 & 0.659 & 0.613 \\
        ENN & 0.540 & 0.500 & 0.100 & 0.185 & 0.111 & 0.132 & 0.711 & 0.556 \\
        Repeated ENN & N/A & N/A & N/A & N/A & N/A & N/A & N/A & N/A \\
        All KNN & 0.571 & 0.636 & 0.215 & 0.176 & 0.109 & 0.119 & 0.696 & 0.556 \\
        Near Miss & 0.487 & 0.044 & 0.879 & 0.086 & -0.018 & 0.038 & 0.438 & 0.444 \\
        Tomek Links & 0.509 & 0.561 & 0.019 & \textbf{0.195} & 0.098 & \textbf{0.137} & \textbf{0.711} & 0.555 \\
        One-Sided Selection & 0.507 & \textbf{0.657} & 0.014 & 0.185 & 0.091 & 0.131 & 0.709 & 0.550 \\
        NCL & N/A & N/A & N/A & N/A & N/A & N/A & N/A & N/A \\
        Cluster Centroids & 0.588 & 0.059 & \textbf{0.950} & 0.126 & 0.073 & 0.077 & 0.626 & 0.624 \\
    \end{tabularx}
    \vspace{1mm}
    \caption{\textbf{Dataset Click Prediction V2.}}
\end{table}

\begin{table}[H]
    \centering
    \setlength\tabcolsep{2pt}
    \begin{tabularx}{\textwidth}{lRRRRRRRR}
        & B. Accuracy & Precision & Recall & F1 Max & MCC & PR AUC & ROC AUC & P-ROC AUC \\
        \midrule
        Baseline & 0.675 & 0.660 & 0.361 & 0.514 & 0.466 & 0.490 & 0.861 & 0.725 \\
        Random Oversampling & 0.702 & 0.578 & 0.424 & 0.517 & 0.468 & 0.478 & 0.863 & \textbf{0.860} \\
        SMOTE & 0.671 & 0.580 & 0.359 & 0.476 & 0.431 & 0.425 & 0.838 & 0.835 \\
        Borderline SMOTE & 0.668 & 0.572 & 0.352 & 0.473 & 0.423 & 0.433 & 0.836 & 0.833 \\
        SVM SMOTE & 0.671 & 0.603 & 0.359 & 0.475 & 0.433 & 0.447 & 0.837 & 0.832 \\
        KMeans SMOTE & N/A & N/A & N/A & N/A & N/A & N/A & N/A & N/A \\
        ADASYN & 0.674 & 0.585 & 0.363 & 0.471 & 0.436 & 0.420 & 0.834 & 0.831 \\
        Random Undersampling & \textbf{0.767} & 0.213 & 0.753 & 0.444 & 0.319 & 0.398 & 0.823 & 0.819 \\
        CNN & 0.766 & 0.366 & 0.643 & 0.475 & 0.412 & 0.452 & 0.838 & 0.819 \\
        ENN & 0.724 & 0.734 & 0.470 & 0.519 & 0.482 & \textbf{0.496} & 0.863 & 0.753 \\
        Repeated ENN & 0.714 & \textbf{0.905} & 0.665 & 0.510 & 0.400 & 0.471 & 0.858 & 0.716 \\
        All KNN & 0.724 & 0.641 & 0.572 & 0.521 & \textbf{0.485} & 0.491 & 0.861 & 0.743 \\
        Near Miss & 0.672 & 0.103 & 0.939 & 0.299 & 0.163 & 0.226 & 0.740 & 0.740 \\
        Tomek Links & 0.581 & 0.765 & 0.165 & 0.517 & 0.340 & 0.478 & 0.861 & 0.725 \\
        One-Sided Selection & 0.688 & 0.650 & 0.388 & 0.519 & 0.480 & 0.486 & 0.861 & 0.728 \\
        NCL & 0.724 & 0.601 & 0.470 & \textbf{0.531} & 0.485 & 0.483 & \textbf{0.864} & 0.751 \\
        Cluster Centroids & 0.749 & 0.167 & \textbf{0.994} & 0.436 & 0.268 & 0.357 & 0.828 & 0.821 \\
    \end{tabularx}
    \vspace{1mm}
    \caption{\textbf{Dataset Amazon Employee.}}
\end{table}

\begin{table}[H]
    \centering
    \setlength\tabcolsep{2pt}
    \begin{tabularx}{\textwidth}{lRRRRRRRR}
        & B. Accuracy & Precision & Recall & F1 Max & MCC & PR AUC & ROC AUC & P-ROC AUC \\
        \midrule
        Baseline & 0.579 & 0.620 & 0.164 & \textbf{0.437} & 0.299 & \textbf{0.396} & \textbf{0.894} & 0.662 \\
        Random Oversampling & N/A & N/A & N/A & N/A & N/A & N/A & N/A & N/A \\
        SMOTE & \textbf{0.802} & 0.259 & 0.759 & 0.426 & 0.378 & 0.358 & 0.875 & \textbf{0.874} \\
        Borderline SMOTE & 0.785 & 0.304 & 0.716 & 0.418 & \textbf{0.382} & 0.322 & 0.870 & 0.868 \\
        SVM SMOTE & N/A & N/A & N/A & N/A & N/A & N/A & N/A & N/A \\
        KMeans SMOTE & N/A & N/A & N/A & N/A & N/A & N/A & N/A & N/A \\
        ADASYN & 0.800 & 0.245 & \textbf{0.810} & 0.425 & 0.369 & 0.348 & 0.873 & 0.871 \\
        Random Undersampling & N/A & N/A & N/A & N/A & N/A & N/A & N/A & N/A \\
        CNN & N/A & N/A & N/A & N/A & N/A & N/A & N/A & N/A \\
        ENN & 0.780 & 0.590 & 0.801 & 0.437 & 0.323 & 0.373 & 0.893 & 0.660 \\
        Repeated ENN & N/A & N/A & N/A & N/A & N/A & N/A & N/A & N/A \\
        All KNN & N/A & N/A & N/A & N/A & N/A & N/A & N/A & N/A \\
        Near Miss & 0.672 & 0.107 & 0.758 & 0.187 & 0.167 & 0.081 & 0.646 & 0.625 \\
        Tomek Links & N/A & N/A & N/A & N/A & N/A & N/A & N/A & N/A \\
        One-Sided Selection & 0.579 & \textbf{0.621} & 0.164 & 0.437 & 0.300 & 0.395 & 0.893 & 0.662 \\
        NCL & 0.603 & 0.542 & 0.218 & 0.435 & 0.318 & 0.371 & 0.893 & 0.658 \\
        Cluster Centroids & N/A & N/A & N/A & N/A & N/A & N/A & N/A & N/A \\
    \end{tabularx}
    \vspace{1mm}
    \caption{\textbf{Dataset BNG - Sick.}}
\end{table}

\begin{table}[H]
    \centering
    \setlength\tabcolsep{2pt}
    \begin{tabularx}{\textwidth}{lRRRRRRRR}
        & B. Accuracy & Precision & Recall & F1 Max & MCC & PR AUC & ROC AUC & P-ROC AUC \\
        \midrule
        Baseline & 0.996 & 0.940 & 0.995 & 0.971 & \textbf{0.965} & 0.986 & 0.999 & 0.995 \\
        Random Oversampling & 0.989 & \textbf{0.956} & 0.982 & 0.969 & 0.959 & 0.984 & 0.999 & 0.999 \\
        SMOTE & 0.989 & 0.940 & 0.982 & 0.969 & 0.958 & 0.984 & 0.999 & 0.999 \\
        Borderline SMOTE & \textbf{0.997} & 0.948 & \textbf{1.000} & 0.973 & 0.962 & 0.987 & 0.999 & 0.999 \\
        SVM SMOTE & 0.997 & 0.936 & \textbf{1.000} & 0.971 & 0.961 & 0.986 & 0.999 & 0.999 \\
        KMeans SMOTE & 0.991 & 0.948 & 0.986 & 0.967 & 0.965 & 0.982 & 0.999 & 0.999 \\
        ADASYN & 0.991 & 0.952 & 0.986 & 0.967 & 0.960 & 0.980 & 0.999 & 0.999 \\
        Random Undersampling & 0.987 & 0.709 & \textbf{1.000} & 0.961 & 0.831 & 0.981 & 0.999 & 0.999 \\
        CNN & N/A & N/A & N/A & N/A & N/A & N/A & N/A & N/A \\
        ENN & 0.997 & 0.936 & \textbf{1.000} & 0.961 & 0.956 & 0.974 & 0.999 & 0.993 \\
        Repeated ENN & 0.995 & 0.932 & 0.995 & 0.963 & 0.961 & 0.977 & 0.999 & 0.993 \\
        All KNN & 0.995 & 0.925 & 0.995 & 0.961 & 0.957 & 0.972 & 0.999 & 0.992 \\
        Near Miss & 0.997 & 0.940 & \textbf{1.000} & \textbf{0.974} & \textbf{0.965} & \textbf{0.990} & \textbf{1.000} & \textbf{1.000} \\
        Tomek Links & 0.994 & 0.909 & 0.995 & 0.961 & 0.948 & 0.978 & 0.999 & 0.993 \\
        One-Sided Selection & 0.991 & 0.932 & 0.986 & 0.959 & 0.956 & 0.978 & 0.999 & 0.992 \\
        NCL & 0.994 & 0.916 & 0.995 & 0.960 & 0.947 & 0.973 & 0.999 & 0.992 \\
        Cluster Centroids & 0.996 & 0.899 & \textbf{1.000} & 0.957 & 0.945 & 0.982 & 0.999 & 0.999 \\
    \end{tabularx}
    \vspace{1mm}
    \caption{\textbf{Dataset Sylva Prior.}}
\end{table}

\begin{table}[H]
    \centering
    \setlength\tabcolsep{2pt}
    \begin{tabularx}{\textwidth}{lRRRRRRRR}
        & B. Accuracy & Precision & Recall & F1 Max & MCC & PR AUC & ROC AUC & P-ROC AUC \\
        \midrule
        Baseline & 0.595 & 0.731 & 0.196 & 0.506 & 0.354 & 0.491 & 0.894 & 0.709 \\
        Random Oversampling & 0.803 & 0.306 & 0.798 & 0.477 & 0.413 & 0.396 & 0.845 & 0.678 \\
        SMOTE & 0.791 & 0.357 & 0.718 & 0.479 & 0.429 & 0.424 & 0.851 & 0.850 \\
        Borderline SMOTE & 0.762 & 0.375 & 0.652 & 0.463 & 0.410 & 0.384 & 0.849 & 0.663 \\
        SVM SMOTE & 0.774 & 0.361 & 0.655 & 0.478 & 0.423 & 0.415 & 0.853 & \textbf{0.850} \\
        KMeans SMOTE & N/A & N/A & N/A & N/A & N/A & N/A & N/A & N/A \\
        ADASYN & 0.790 & 0.347 & 0.738 & 0.475 & 0.422 & 0.407 & 0.849 & 0.679 \\
        Random Undersampling & \textbf{0.826} & 0.333 & 0.840 & 0.505 & \textbf{0.450} & \textbf{0.495} & \textbf{0.895} & 0.709 \\
        CNN & N/A & N/A & N/A & N/A & N/A & N/A & N/A & N/A \\
        ENN & 0.811 & 0.580 & 0.774 & 0.506 & 0.441 & 0.459 & 0.893 & 0.701 \\
        Repeated ENN & N/A & N/A & N/A & N/A & N/A & N/A & N/A & N/A \\
        All KNN & 0.811 & 0.449 & \textbf{0.900} & 0.498 & 0.408 & 0.432 & 0.889 & 0.727 \\
        Near Miss & 0.751 & 0.254 & 0.802 & 0.416 & 0.332 & 0.294 & 0.839 & 0.833 \\
        Tomek Links & 0.594 & 0.733 & 0.195 & \textbf{0.507} & 0.354 & 0.492 & 0.894 & 0.709 \\
        One-Sided Selection & 0.594 & \textbf{0.734} & 0.195 & 0.506 & 0.353 & 0.492 & 0.894 & 0.709 \\
        NCL & 0.686 & 0.514 & 0.414 & 0.506 & 0.394 & 0.489 & 0.893 & 0.721 \\
        Cluster Centroids & N/A & N/A & N/A & N/A & N/A & N/A & N/A & N/A \\
    \end{tabularx}
    \vspace{1mm}
    \caption{\textbf{Dataset BNG - Spect.}}
\end{table}

\begin{table}[H]
    \centering
    \setlength\tabcolsep{2pt}
    \begin{tabularx}{\textwidth}{lRRRRRRRR}
        & B. Accuracy & Precision & Recall & F1 Max & MCC & PR AUC & ROC AUC & P-ROC AUC \\
        \midrule
        Baseline & 0.994 & \textbf{0.979} & 0.988 & 0.984 & 0.983 & \textbf{0.999} & \textbf{1.000} & 0.999 \\
        Random Oversampling & 0.995 & 0.976 & 0.991 & 0.985 & 0.983 & 0.998 & 1.000 & \textbf{1.000} \\
        SMOTE & 0.997 & 0.968 & 0.994 & 0.981 & 0.980 & 0.998 & 1.000 & 1.000 \\
        Borderline SMOTE & 0.997 & 0.969 & 0.994 & 0.981 & 0.980 & 0.997 & 1.000 & 1.000 \\
        SVM SMOTE & 0.998 & 0.968 & 0.997 & 0.983 & 0.980 & 0.998 & 1.000 & 1.000 \\
        KMeans SMOTE & N/A & N/A & N/A & N/A & N/A & N/A & N/A & N/A \\
        ADASYN & 0.998 & 0.963 & 0.996 & 0.982 & 0.979 & 0.997 & 1.000 & 1.000 \\
        Random Undersampling & 0.998 & 0.893 & \textbf{1.000} & 0.980 & 0.943 & 0.995 & 1.000 & 1.000 \\
        CNN & N/A & N/A & N/A & N/A & N/A & N/A & N/A & N/A \\
        ENN & \textbf{0.998} & 0.975 & 0.998 & 0.984 & 0.983 & 0.997 & 1.000 & 0.998 \\
        Repeated ENN & N/A & N/A & N/A & N/A & N/A & N/A & N/A & N/A \\
        All KNN & 0.997 & 0.969 & 0.996 & 0.983 & 0.982 & 0.989 & 1.000 & 0.996 \\
        Near Miss & 0.994 & 0.864 & 0.996 & 0.959 & 0.924 & 0.991 & 1.000 & 1.000 \\
        Tomek Links & 0.995 & 0.974 & 0.991 & 0.983 & 0.982 & 0.998 & 1.000 & 0.999 \\
        One-Sided Selection & 0.995 & 0.973 & 0.990 & 0.985 & 0.981 & 0.998 & 1.000 & 0.999 \\
        NCL & 0.998 & 0.974 & 0.996 & \textbf{0.985} & \textbf{0.983} & 0.996 & 1.000 & 0.998 \\
        Cluster Centroids & 0.970 & 0.291 & 0.999 & 0.842 & 0.522 & 0.914 & 0.998 & 0.998 \\
    \end{tabularx}
    \vspace{1mm}
    \caption{\textbf{Dataset CIC-IDS-2017.}}
\end{table}

\begin{table}[H]
    \centering
    \setlength\tabcolsep{2pt}
    \begin{tabularx}{\textwidth}{lRRRRRRRR}
        & B. Accuracy & Precision & Recall & F1 Max & MCC & PR AUC & ROC AUC & P-ROC AUC \\
        \midrule
        Baseline & 0.945 & \textbf{0.966} & 0.891 & 0.929 & 0.924 & 0.982 & 0.998 & 0.988 \\
        Random Oversampling & 0.952 & 0.888 & 0.910 & 0.908 & 0.893 & 0.851 & 0.953 & 0.953 \\
        SMOTE & 0.950 & 0.964 & 0.902 & 0.933 & \textbf{0.929} & 0.981 & 0.998 & 0.998 \\
        Borderline SMOTE & 0.951 & 0.963 & 0.905 & \textbf{0.934} & 0.928 & 0.982 & 0.998 & 0.998 \\
        SVM SMOTE & 0.950 & 0.962 & 0.902 & 0.931 & 0.927 & 0.982 & 0.999 & \textbf{0.999} \\
        KMeans SMOTE & N/A & N/A & N/A & N/A & N/A & N/A & N/A & N/A \\
        ADASYN & 0.948 & 0.959 & 0.898 & 0.930 & 0.924 & 0.982 & 0.998 & 0.998 \\
        Random Undersampling & \textbf{0.978} & 0.716 & 0.985 & 0.902 & 0.827 & 0.972 & 0.998 & 0.998 \\
        CNN & N/A & N/A & N/A & N/A & N/A & N/A & N/A & N/A \\
        ENN & 0.954 & 0.963 & 0.914 & 0.925 & 0.920 & 0.982 & 0.998 & 0.988 \\
        Repeated ENN & 0.970 & 0.961 & 0.948 & 0.929 & 0.924 & 0.983 & \textbf{0.999} & 0.989 \\
        All KNN & 0.962 & 0.957 & 0.933 & 0.930 & 0.924 & \textbf{0.983} & 0.999 & 0.989 \\
        Near Miss & 0.813 & 0.134 & \textbf{0.993} & 0.777 & 0.288 & 0.720 & 0.948 & 0.948 \\
        Tomek Links & 0.944 & 0.963 & 0.889 & 0.929 & 0.921 & 0.983 & 0.999 & 0.989 \\
        One-Sided Selection & 0.943 & 0.960 & 0.889 & 0.924 & 0.920 & 0.982 & 0.999 & 0.989 \\
        NCL & 0.959 & 0.940 & 0.923 & 0.925 & 0.919 & 0.981 & 0.998 & 0.989 \\
        Cluster Centroids & N/A & N/A & N/A & N/A & N/A & N/A & N/A & N/A \\
    \end{tabularx}
    \vspace{1mm}
    \caption{\textbf{Dataset UNSW-NB15.}}
\end{table}

\begin{table}[H]
    \centering
    \setlength\tabcolsep{2pt}
    \begin{tabularx}{\textwidth}{lRRRRRRRR}
        & B. Accuracy & Precision & Recall & F1 Max & MCC & PR AUC & ROC AUC & P-ROC AUC \\
        \midrule
        Baseline & 0.981 & 0.978 & 0.964 & 0.978 & 0.967 & \textbf{0.993} & \textbf{0.999} & 0.995 \\
        Random Oversampling & 0.980 & 0.978 & 0.964 & 0.974 & 0.963 & 0.988 & 0.997 & 0.997 \\
        SMOTE & 0.981 & 0.985 & 0.964 & 0.978 & 0.971 & 0.992 & 0.998 & \textbf{0.998} \\
        Borderline SMOTE & \textbf{0.985} & \textbf{0.985} & 0.971 & 0.978 & \textbf{0.976} & 0.991 & 0.998 & 0.998 \\
        SVM SMOTE & \textbf{0.985} & \textbf{0.985} & 0.971 & \textbf{0.985} & \textbf{0.976} & 0.991 & 0.998 & 0.998 \\
        KMeans SMOTE & N/A & N/A & N/A & N/A & N/A & N/A & N/A & N/A \\
        ADASYN & 0.981 & 0.985 & 0.964 & 0.978 & 0.971 & 0.989 & 0.998 & 0.998 \\
        Random Undersampling & 0.977 & 0.924 & 0.964 & 0.961 & 0.937 & 0.980 & 0.996 & 0.996 \\
        CNN & 0.978 & 0.937 & 0.964 & 0.967 & 0.944 & 0.988 & 0.997 & 0.997 \\
        ENN & 0.980 & 0.978 & 0.964 & 0.975 & 0.963 & 0.990 & 0.998 & 0.991 \\
        Repeated ENN & 0.981 & 0.978 & 0.964 & 0.978 & 0.967 & 0.991 & 0.998 & 0.992 \\
        All KNN & 0.982 & 0.971 & 0.971 & 0.971 & 0.963 & 0.988 & 0.997 & 0.991 \\
        Near Miss & 0.980 & 0.950 & \textbf{1.000} & 0.971 & 0.952 & 0.989 & 0.998 & 0.998 \\
        Tomek Links & 0.981 & 0.978 & 0.964 & 0.971 & 0.967 & 0.990 & 0.997 & 0.991 \\
        One-Sided Selection & 0.981 & 0.985 & 0.964 & 0.978 & 0.971 & 0.989 & 0.997 & 0.991 \\
        NCL & 0.983 & 0.978 & 0.971 & 0.974 & 0.967 & 0.986 & 0.996 & 0.991 \\
        Cluster Centroids & 0.978 & 0.905 & 0.986 & 0.964 & 0.922 & 0.984 & 0.997 & 0.997 \\
    \end{tabularx}
    \vspace{1mm}
    \caption{\textbf{Dataset CIC-Evasive-PDF.}}
\end{table}

\begin{table}[H]
    \centering
    \setlength\tabcolsep{2pt}
    \begin{tabularx}{\textwidth}{lRRRRRRRR}
        & B. Accuracy & Precision & Recall & F1 Max & MCC & PR AUC & ROC AUC & P-ROC AUC \\
        \midrule
        Baseline & 0.937 & \textbf{0.980} & 0.877 & \textbf{0.931} & \textbf{0.918} & \textbf{0.974} & \textbf{0.993} & 0.981 \\
        Random Oversampling & 0.902 & 0.797 & 0.834 & 0.819 & 0.786 & 0.715 & 0.907 & 0.907 \\
        SMOTE & 0.917 & 0.774 & 0.871 & 0.833 & 0.792 & 0.749 & 0.934 & 0.934 \\
        Borderline SMOTE & 0.919 & 0.791 & 0.870 & 0.835 & 0.798 & 0.759 & 0.936 & 0.936 \\
        SVM SMOTE & 0.920 & 0.779 & 0.876 & 0.837 & 0.799 & 0.756 & 0.938 & 0.938 \\
        KMeans SMOTE & N/A & N/A & N/A & N/A & N/A & N/A & N/A & N/A \\
        ADASYN & 0.919 & 0.777 & 0.871 & 0.837 & 0.797 & 0.755 & 0.936 & 0.935 \\
        Random Undersampling & \textbf{0.956} & 0.809 & 0.948 & 0.908 & 0.854 & 0.964 & 0.990 & \textbf{0.990} \\
        CNN & N/A & N/A & N/A & N/A & N/A & N/A & N/A & N/A \\
        ENN & 0.949 & 0.973 & 0.911 & 0.926 & 0.915 & 0.972 & 0.992 & 0.980 \\
        Repeated ENN & 0.953 & 0.973 & 0.933 & 0.928 & 0.917 & 0.972 & 0.992 & 0.980 \\
        All KNN & 0.951 & 0.931 & 0.926 & 0.920 & 0.907 & 0.970 & 0.992 & 0.980 \\
        Near Miss & 0.947 & 0.827 & \textbf{0.981} & 0.906 & 0.855 & 0.957 & 0.988 & 0.987 \\
        Tomek Links & 0.894 & 0.812 & 0.814 & 0.819 & 0.788 & 0.756 & 0.913 & 0.891 \\
        One-Sided Selection & 0.937 & 0.977 & 0.877 & 0.928 & 0.916 & 0.972 & 0.992 & 0.979 \\
        NCL & 0.950 & 0.922 & 0.911 & 0.920 & 0.905 & 0.969 & 0.992 & 0.981 \\
        Cluster Centroids & N/A & N/A & N/A & N/A & N/A & N/A & N/A & N/A \\
    \end{tabularx}
    \vspace{1mm}
    \caption{\textbf{Dataset Ember.}}
\end{table}

\begin{table}[H]
    \centering
    \setlength\tabcolsep{2pt}
    \begin{tabularx}{\textwidth}{lRRRRRRRR}
        & B. Accuracy & Precision & Recall & F1 Max & MCC & PR AUC & ROC AUC & P-ROC AUC \\
        \midrule
        Baseline & 0.816 & 0.833 & 0.686 & 0.769 & 0.674 & 0.814 & 0.923 & 0.880 \\
        Random Oversampling & \textbf{0.866} & 0.778 & 0.824 & 0.811 & \textbf{0.719} & 0.845 & 0.932 & 0.927 \\
        SMOTE & \textbf{0.866} & 0.778 & 0.843 & 0.808 & \textbf{0.719} & \textbf{0.871} & 0.934 & 0.929 \\
        Borderline SMOTE & 0.842 & 0.833 & 0.804 & 0.809 & 0.677 & 0.822 & 0.925 & 0.925 \\
        SVM SMOTE & 0.852 & 0.792 & 0.824 & \textbf{0.816} & 0.692 & 0.865 & 0.935 & 0.927 \\
        KMeans SMOTE & 0.862 & 0.822 & 0.824 & 0.804 & 0.708 & 0.867 & \textbf{0.938} & \textbf{0.934} \\
        ADASYN & 0.843 & 0.677 & 0.863 & 0.808 & 0.639 & 0.835 & 0.925 & 0.923 \\
        Random Undersampling & 0.844 & 0.792 & 0.804 & 0.779 & 0.681 & 0.784 & 0.926 & 0.916 \\
        CNN & 0.794 & 0.673 & \textbf{1.000} & 0.769 & 0.574 & 0.787 & 0.926 & 0.883 \\
        ENN & 0.836 & 0.841 & 0.725 & 0.779 & 0.704 & 0.814 & 0.927 & 0.894 \\
        Repeated ENN & 0.836 & 0.841 & 0.961 & 0.779 & 0.704 & 0.824 & 0.927 & 0.891 \\
        All KNN & 0.830 & \textbf{0.854} & 0.961 & 0.779 & 0.688 & 0.800 & 0.927 & 0.892 \\
        Near Miss & 0.845 & 0.833 & 0.922 & 0.792 & 0.674 & 0.797 & 0.931 & 0.931 \\
        Tomek Links & 0.822 & 0.818 & 0.706 & 0.766 & 0.676 & 0.815 & 0.925 & 0.884 \\
        One-Sided Selection & 0.822 & 0.818 & 0.706 & 0.771 & 0.676 & 0.805 & 0.923 & 0.881 \\
        NCL & 0.823 & 0.795 & 0.824 & 0.769 & 0.653 & 0.813 & 0.927 & 0.892 \\
        Cluster Centroids & 0.844 & 0.837 & 0.765 & 0.788 & 0.696 & 0.825 & 0.932 & 0.930 \\
    \end{tabularx}
    \vspace{1mm}
    \caption{\textbf{Dataset Graph - Embedding.}}
\end{table}

\begin{table}[H]
    \centering
    \setlength\tabcolsep{2pt}
    \begin{tabularx}{\textwidth}{lRRRRRRRR}
        & B. Accuracy & Precision & Recall & F1 Max & MCC & PR AUC & ROC AUC & P-ROC AUC \\
        \midrule
        Baseline & 0.773 & 0.795 & 0.608 & 0.776 & 0.598 & 0.834 & 0.903 & 0.868 \\
        Random Oversampling & 0.828 & 0.814 & 0.725 & 0.796 & 0.678 & 0.840 & 0.907 & 0.907 \\
        SMOTE & 0.842 & 0.795 & 0.784 & 0.796 & 0.681 & 0.861 & 0.920 & 0.914 \\
        Borderline SMOTE & 0.822 & 0.818 & 0.706 & 0.800 & 0.676 & 0.846 & 0.908 & 0.904 \\
        SVM SMOTE & 0.848 & 0.814 & 0.804 & \textbf{0.816} & 0.681 & 0.863 & 0.915 & 0.915 \\
        KMeans SMOTE & 0.828 & 0.804 & 0.725 & 0.804 & 0.678 & 0.852 & \textbf{0.925} & \textbf{0.920} \\
        ADASYN & \textbf{0.858} & 0.810 & 0.824 & 0.800 & \textbf{0.697} & 0.860 & 0.922 & 0.917 \\
        Random Undersampling & 0.846 & 0.776 & 0.784 & 0.777 & 0.688 & 0.842 & 0.905 & 0.905 \\
        CNN & 0.795 & 0.522 & \textbf{0.922} & 0.771 & 0.532 & 0.826 & 0.903 & 0.901 \\
        ENN & 0.839 & 0.805 & 0.882 & 0.800 & 0.645 & 0.837 & 0.909 & 0.894 \\
        Repeated ENN & 0.847 & 0.821 & 0.882 & 0.778 & 0.648 & 0.833 & 0.903 & 0.882 \\
        All KNN & 0.827 & 0.795 & 0.843 & 0.750 & 0.647 & 0.834 & 0.903 & 0.872 \\
        Near Miss & 0.811 & 0.762 & 0.843 & 0.740 & 0.618 & 0.809 & 0.890 & 0.890 \\
        Tomek Links & 0.797 & 0.825 & 0.647 & 0.757 & 0.643 & 0.840 & 0.906 & 0.873 \\
        One-Sided Selection & 0.793 & 0.805 & 0.647 & 0.757 & 0.629 & 0.832 & 0.898 & 0.867 \\
        NCL & 0.843 & 0.800 & 0.843 & 0.780 & 0.655 & 0.839 & 0.904 & 0.881 \\
        Cluster Centroids & 0.834 & \textbf{0.903} & 0.784 & 0.792 & 0.669 & \textbf{0.868} & \textbf{0.925} & 0.920 \\
    \end{tabularx}
    \vspace{1mm}
    \caption{\textbf{Dataset Graph - Raw.}}
\end{table}

\end{document}